\documentclass[sigconf,nonacm]{acmart}

\setcopyright{none}
\renewcommand\footnotetextcopyrightpermission[1]{}
\usepackage{wasysym}        
\usepackage{multirow}
\usepackage{makecell}
\usepackage{tikz}
\usetikzlibrary{positioning,arrows.meta,fit,backgrounds,calc,shapes.geometric}

\newcommand{\full}{\CIRCLE}
\newcommand{\half}{\LEFTcircle}
\newcommand{\none}{\Circle}

\newcommand{\Cop}{\mathcal{C}}
\newcommand{\Uop}{\mathcal{U}}
\newcommand{\Hist}{H}
\newcommand{\Zrep}{Z}
\newcommand{\budget}{B}
\newcommand{\dist}{\mathcal{D}}
\newcommand{\loss}{\ell}
\newcommand{\E}{\mathbb{E}}
\newcommand{\I}{\mathrm{I}}
\newcommand{\rate}{\mathrm{rate}}
\newcommand{\Istar}{I^\star}

\begin{document}

\title{What to Keep, What to Forget: A Rate--Distortion View of Memory Compaction in LLMs and Agents}

\author{Ashwin Gerard Colaco}
\email{acolaco@uci.edu}
\affiliation{%
  \institution{University of California, Irvine}
  \city{Irvine}
  \state{California}
  \country{USA}
}

\author{Nada Lahjouji}
\email{nlahjouj@uci.edu}
\affiliation{%
  \institution{University of California, Irvine}
  \city{Irvine}
  \state{California}
  \country{USA}
}

\renewcommand{\shortauthors}{Colaco and Lahjouji}

\begin{abstract}
Large language models, and the agents built on them, spend an ever-growing share
of their compute and memory on remembering: caching attention keys and
values, carrying long prompts, maintaining recurrent state, and storing what
happened in previous turns and sessions. Because none of this memory is free,
four largely separate research communities have each learned to compact
it. They evict or quantize the KV cache, prune or distill prompts, bound
architectural state, and consolidate agent memory. We argue that these are
instances of one problem: a rate--distortion decision about \textbf{what
context-derived information to retain versus discard, at what fidelity, under a
resource budget}, so as to preserve downstream task utility. We make this lens
precise with a single compaction objective and a layer-agnostic lower bound, use
it to build a seven-axis taxonomy that classifies methods from across the stack
uniformly, and use it to transfer mechanisms between layers that have never
been connected, from serving-stack KV management to agent long-term memory. Two
patterns hold across the survey. At every layer the signal that decides what to
keep is attention magnitude or recency, and it fails in the same way everywhere, by
discarding, before the query is known and with no way to undo it, information the
query later needs. And while compression is measured carefully on single-turn long
context, the repeated compaction that agents actually perform is almost never
measured, and no benchmark holds one budget axis across all the layers at once. We
turn both observations into a benchmark proposal, a small reference experiment, and
a set of compaction-aware design principles, and we map the open problems.
\end{abstract}



\maketitle

\section{Introduction}
\label{sec:intro}

Run a transformer over a hundred thousand tokens and the cached keys and values of
everything it has already read take up more of the accelerator than the model's
own weights~\citep{pagedattention,kivi}. The same pressure appears in other forms.
An agent working toward a goal builds up a record of tool calls and their results
that soon outgrows any context window~\citep{react,memgpt}; a state-space or
linear-attention model squeezes the whole past into a state of fixed size; an
assistant expected to know its user has to carry something from one session into
the next~\citep{locomo,longmemeval}. In each case the scarce resource is no longer
computation but the memory of what has been read, and the practical question is
what to do when there is more of it than fits.

Four bodies of work have grown up around that question with little contact between
them. \textbf{KV-cache compression} acts during inference: it drops low-value
tokens~\citep{h2o,snapkv,streamingllm}, stores the rest at a few
bits~\citep{kivi,kvquant}, factorizes them~\citep{palu,mla}, or merges
them~\citep{cam,minicache}. \textbf{Prompt and context compression} acts earlier,
before the model reads the input, deleting or rewriting tokens~\citep{llmlingua,selectivecontext}
or swapping spans for a few learned ``gist''
vectors~\citep{gisting,icae,xrag}. A third line changes the
\textbf{architecture} so that memory is bounded by construction, through segment
recurrence~\citep{rmt,compressivetransformer}, a fixed compressive
memory~\citep{infiniattention}, or linear-time state-space
models~\citep{mamba,rwkv,titans}. A fourth works at the scale of a whole
\textbf{agent}, trimming the working context during a
task~\citep{acon,contextfolding,memone} and consolidating what is worth keeping
across tasks~\citep{mem0,amem,generativeagents}. Each line carries its own
benchmarks and its own idea of success, whether that is compression ratio at fixed
perplexity, tokens saved at fixed accuracy, or recall at fixed state size, and a
method from one line seldom cites a method from another.

\begin{figure*}[t]
\centering
\footnotesize
\begin{tikzpicture}[
  >=Latex, node distance=2mm,
  root/.style={draw, rounded corners, fill=violet!18, align=center, font=\small\bfseries,
               minimum height=12mm, text width=2.5cm},
  cat/.style={draw, rounded corners, align=center, font=\bfseries, text width=2.55cm, minimum height=7mm},
  leaf/.style={draw, rounded corners, align=left, font=\scriptsize, text width=4.4cm, minimum height=5mm, inner sep=2pt},
  kv/.style={cat, fill=blue!18}, pr/.style={cat, fill=teal!18},
  ar/.style={cat, fill=orange!20}, ag/.style={cat, fill=red!16}, cc/.style={cat, fill=gray!18},
]
\node[root] (root) {Memory\\ compaction\\[1pt]\normalfont\scriptsize one rate--distortion decision};

\node[kv, right=10mm of root, yshift=24mm] (kv) {KV-cache (\S\ref{sec:kv})};
\node[pr, below=of kv]  (pr) {Prompt/context (\S\ref{sec:prompt})};
\node[ar, below=of pr]  (ar) {Architectural (\S\ref{sec:arch})};
\node[ag, below=of ar]  (ag) {Agent memory (\S\ref{sec:agent})};
\node[cc, below=of ag]  (cc) {Cross-cutting (\S\ref{sec:sparse},\,\S\ref{sec:multi})};

\node[leaf, right=8mm of kv] (kvl) {evict $\cdot$ quantize $\cdot$ low-rank $\cdot$ merge\\ \emph{H2O, SnapKV, KIVI, Palu}};
\node[leaf, right=8mm of pr] (prl) {hard prune $\cdot$ soft gist $\cdot$ distill\\ \emph{LLMLingua, gisting, ICAE}};
\node[leaf, right=8mm of ar] (arl) {recurrence $\cdot$ SSM $\cdot$ assoc.\ memory\\ \emph{RMT, Mamba, Titans}};
\node[leaf, right=8mm of ag] (agl) {curate $\cdot$ consolidate $\cdot$ retrieve\\ \emph{MemGPT, Mem0, RAPTOR}};
\node[leaf, right=8mm of cc] (ccl) {learned sparsity $\cdot$ visual/multi-agent\\ \emph{NSA, MoBA, ToMe, G-Memory}};

\foreach \c in {kv,pr,ar,ag,cc}{ \draw[->] (root.east) to[out=0,in=180] (\c.west); }
\foreach \c in {kv,pr,ar,ag,cc}{ \draw[->] (\c.east) -- (\c l.west); }
\end{tikzpicture}
\caption{Map of the survey. The four areas of memory compaction act on different
substrates and at different times, yet each makes the same rate--distortion choice
(Section~\ref{sec:formalism}): the KV cache (Section~\ref{sec:kv}), the
prompt and context (Section~\ref{sec:prompt}), architectural state
(Section~\ref{sec:arch}), and agent memory (Section~\ref{sec:agent}), with
trainable sparse attention (Section~\ref{sec:sparse}) and multimodal or
multi-agent settings (Section~\ref{sec:multi}) cutting across them. The taxonomy of
Section~\ref{sec:taxonomy} classifies the methods on shared axes, and
Section~\ref{sec:bridge} carries mechanisms between them.}
\label{fig:org}
\end{figure*}
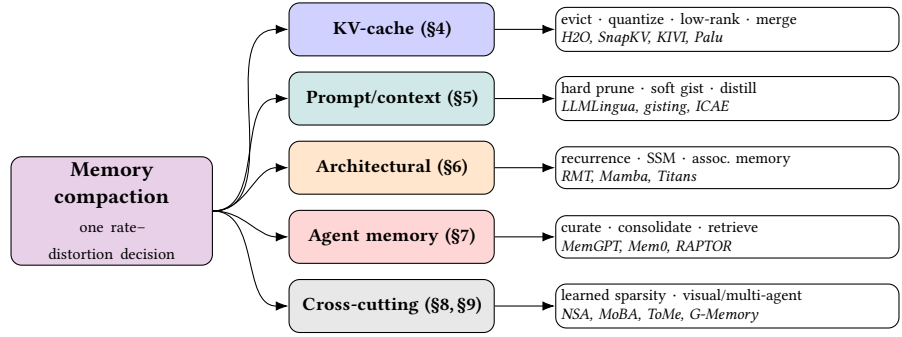

These four are the same problem in different clothes. A method at any level takes
the history $\Hist$ of everything the model has read and produces a smaller
representation $\Zrep$ that fits a budget $\budget$; call this the \emph{compaction
operator} $\Cop$. The model then runs on $\Zrep$ in place of the original, through a
\emph{usage operator} $\Uop$. Stated or not, the method is solving one
optimization: keep $\Zrep$ small while giving up as little downstream accuracy as
possible. That is a rate--distortion problem (Section~\ref{sec:formalism}), and once
the field is read this way the differences between methods come down to three. What
unit do they compress: a bit, a token, an attention head, a span of text, a vector,
a stored fact? Where in the model's life do they act: pretraining, prefill,
decoding, serving, inside a task, or between tasks? And do they decide what to keep
before the query is known, or after? Figure~\ref{fig:org} maps the survey onto these
distinctions.

Reading the field this way buys three things. The first is a shared yardstick. A KV
evictor that keeps a quarter of the tokens and a quantizer that stores every token
at four bits land at the same point on a bytes-per-token axis, so they can be
compared on the accuracy each buys there, and an agent's summary can be placed on
that axis too. The second is that one failure mode turns out to lie behind many
symptoms. A method that fixes what it keeps before the query arrives, and cannot
take the choice back, will sooner or later drop something the query needed, whether
it is evicting KV entries, summarizing a transcript, or folding the past into a
recurrent state. The two properties at work, reversibility and query-conditioning,
recur throughout. The third is that techniques become portable: once an agent's
Ebbinghaus-style forgetting curve~\citep{memorybank} and a KV evictor's
attention-score rule~\citep{h2o} are seen as two estimates of the same quantity, the
better-studied one is a design for the other, the no-eviction retrieval of
Quest~\citep{quest} suggests an agent-memory design, and the output-error bound of
Ada-KV~\citep{adakv} suggests a stopping rule for summarization
(Section~\ref{sec:bridge}).

We are not the first to connect these areas. One recent survey gathers token-level,
parametric, latent, and agent memory under a Forms/Functions/Dynamics
scheme~\citep{surveyagentmem}, and another casts agent memory as a
write--manage--read POMDP~\citep{surveyagentmem2}. What this survey adds is a single
optimization objective with a provable bound rather than a descriptive grouping;
coverage that reaches the serving stack and the quantization and low-rank
mathematics alongside agent memory, which no prior survey spans; the explicit
transfer of mechanisms between layers; and a benchmark we both define and run.
Table~\ref{tab:coverage} sets out the comparison.

We make six contributions.
\begin{itemize}\itemsep2pt
\item We formalize memory compaction as one rate--distortion problem and derive a
  layer-agnostic lower bound (Eq.~\ref{eq:bound}) showing that, below a
  task's information requirement, every layer must err at a rate set by the same
  expression, and that query-agnostic compaction pays a quantifiable penalty
  (Section~\ref{sec:formalism}).
\item We give a seven-axis taxonomy (Section~\ref{sec:taxonomy}) that
  classifies methods from all four layers uniformly, and a master table.
\item We survey each layer through the formalism: KV cache
  (Section~\ref{sec:kv}), prompt/context (Section~\ref{sec:prompt}), architectural
  (Section~\ref{sec:arch}), agent/semantic (Section~\ref{sec:agent}), the trainable
  sparse-attention frontier (Section~\ref{sec:sparse}), and multimodal/multi-agent
  settings (Section~\ref{sec:multi}).
\item We make the inference$\leftrightarrow$agent-memory bridge explicit
  (Section~\ref{sec:bridge}), porting concrete mechanisms across layers and
  deriving five compaction-aware design principles.
\item We propose \textbf{COMPACT-Bench} and run a small reference experiment on
  commodity hardware that places KV eviction, quantization, prompt compression,
  and summarization on one budget axis and measures error accumulation under
  repeated compaction (Sections~\ref{sec:eval}--\ref{sec:exp}).
\item We consolidate the open problems into a prioritized agenda
  (Section~\ref{sec:open}).
\end{itemize}

Two themes return throughout. One is that reversibility usually matters more than
any scoring trick: at the same budget, a method that can fetch back what it
discarded beats one that cannot, and it wins precisely on the queries that turn on
the discarded material. The other is a measurement gap. Compaction is tested on
single-turn long-context tasks, but agents compact the same memory again and again,
and almost nothing measures what that repetition costs; Section~\ref{sec:eval}
proposes a benchmark that does.

We restrict attention to the compaction of memory that comes from a model's
context, drawing mostly on work from 2023 to 2026 in the machine learning, natural
language processing, and systems literatures, preprints included. A method is in
scope if it shrinks the memory or token footprint of the context while trying to
keep the model's behavior intact. Weight-level compression (pruning, distillation,
weight quantization) is out of scope, as is long-context training that imposes no
memory bound, except where such work builds in a bounded memory. Much of the
literature is recent and unrefereed, so we treat preprint numbers as provisional and
flag results that have not been reproduced.

\begin{table*}[t]
\centering
\small
\setlength{\tabcolsep}{5pt}
\renewcommand{\arraystretch}{1.18}
\begin{tabular}{l ccccccc c}
\toprule
\textbf{Survey} & \textbf{KV} & \textbf{Prompt} & \textbf{Arch.} & \textbf{Agent} & \textbf{Sys.} & \makecell{\textbf{RD/IB}\\\textbf{objective}} & \makecell{\textbf{X-layer}\\\textbf{transfer}} & \makecell{\textbf{Unified}\\\textbf{benchmark}} \\
\midrule
Memory in the Age of AI Agents~\cite{surveyagentmem}      & \half & \half & \half & \full & \none & \none & \half & \none \\
KV Cache Management~\cite{surveykvmgmt}                    & \full & \none & \none & \none & \full & \none & \none & \none \\
KV Cache Compression: A Review~\cite{surveykvreview}      & \full & \none & \none & \none & \half & \none & \none & \none \\
Prompt Compression~\cite{surveypromptcomp}               & \none & \full & \none & \none & \none & \half & \none & \none \\
Memory for Autonomous Agents~\cite{surveyagentmem2}      & \none & \half & \half & \full & \none & \half & \half & \half \\
Security of Agent Memory~\cite{surveyagentsec}           & \none & \none & \none & \half & \none & \none & \none & \none \\
Efficient LLMs~\cite{surveyefficientllm}                 & \half & \none & \none & \none & \full & \none & \none & \none \\
\midrule
\textbf{This survey}                                      & \full & \full & \full & \full & \full & \full & \full & \full \\
\bottomrule
\end{tabular}
\caption{Coverage of this survey relative to the closest prior surveys.
\full~dedicated treatment; \half~discussed within another topic; \none~absent.
Columns are the four compaction layers (KV cache, prompt/context,
architectural/recurrent, agent/semantic), the serving/systems substrate, a
unifying \emph{rate--distortion / information-bottleneck objective}, explicit
\emph{cross-layer mechanism transfer}, and a \emph{unified compaction benchmark}.
Recent surveys span the first four columns (notably~\citealp{surveyagentmem});
our novelty is the last three.}
\label{tab:coverage}
\end{table*}

\section{A Unified Formalism for Compaction}
\label{sec:formalism}

This section turns the informal claim into something we can compute with: an
objective, a lower bound that holds across all four layers, a reading of the field's
``importance'' heuristics as estimates of one quantity, and the two properties that
most often decide whether a method fails.

Fix a point during generation, and let $\Hist$ stand for everything the model can
draw on there: the prompt tokens and their
activations, the KV cache, any recurrent state, and any memory of prior turns. A
\emph{compaction scheme} is a pair $(\Cop_\theta, \Uop)$. The compaction operator
$\Cop_\theta : \Hist \mapsto \Zrep$ produces a compact representation $\Zrep$ whose
size obeys a budget, $\rate(\Zrep) \le \budget$, where $\rate(\cdot)$ measures
memory in a currency appropriate to the layer (GPU bytes for a KV cache, tokens
for a prompt, state dimensions for a recurrent model, store size for an agent).
The usage operator $\Uop$ consumes $\Zrep$, almost always by continuing the
model's computation, to produce an output $\hat Y$. For a task instance with
query $Q$ and target $Y$, the scheme's quality is its expected loss, and the
design problem is
\begin{equation}
\label{eq:obj}
\min_{\theta}\ \underbrace{\E_{(\Hist,Q,Y)}\!\big[\loss(\Uop(\Cop_\theta(\Hist),Q),\,Y)\big]}_{\textstyle \dist(\theta)\ =\ \text{distortion}}
\quad \text{s.t.}\quad \rate(\Zrep) \le \budget .
\end{equation}
Equation~\eqref{eq:obj} is a rate--distortion problem: trade memory rate against
task distortion. Its information-bottleneck form makes the trade explicit, namely
maximize $\I(\Zrep; Y \mid Q)$ subject to $\I(\Zrep; \Hist) \le \budget$%
~\citep{tishbyib,quitox}: keep the bits of the history that predict the answer,
spend nothing on the rest.

The objective is not vacuous, and its
consequences are the same at every layer. Let $\Istar(Q) = \I(Y; \Hist \mid Q)$ be
the \emph{task-conditioned information content}: the number of bits of $\Hist$
genuinely required to answer $Q$. Because $\hat Y = \Uop(\Zrep, Q)$ is a function
of $(\Zrep, Q)$ and, for a query-agnostic $\Cop$, the chain
$Y \!-\! \Hist \!-\! \Zrep$ is Markov given $Q$, the data-processing inequality
gives $\I(Y; \hat Y \mid Q) \le \min(\Istar(Q), \budget)$. Fano's inequality then
bounds the error probability $P_e = \Pr[\hat Y \neq Y]$ over an answer space
$\mathcal{Y}$:
\begin{equation}
\label{eq:bound}
P_e \;\ge\; \frac{H(Y \mid Q) - \budget - 1}{\log|\mathcal{Y}|},
\qquad \text{whenever } \budget < \Istar(Q).
\end{equation}
The bound holds for any compact memory $\Zrep$, whether a KV cache, a gist vector,
a recurrent state, or an agent's note store, which is what lets it unify the
field. Three consequences follow directly. (i)~Below the task's information
requirement $\Istar(Q)$, every layer must make errors, at a rate governed
by the same expression; there is no architecture-specific escape. (ii)~The
compression ratio achievable at fixed quality is task-dependent,
$\propto 1/\Istar(Q)$: tasks whose answers carry many bits (exact or multi-hop
retrieval) compress poorly, while tasks whose answers are low-entropy
(classification, gisting of redundant prose) compress well. This is exactly why
the same KV evictor that is near-lossless on summarization collapses on
retrieval. (iii)~A query-agnostic operator must spread its budget across the whole
query distribution, so its effective per-query budget is roughly $\budget - H(Q)$,
whereas a query-conditioned operator can spend all of $\budget$ on the
answer-relevant bits; the gap $H(Q)$ is the precise price of not knowing the
query. The $\Theta(nd)$ lower bound on exact-attention
memory~\citep{compressionbarriers} and the impossibility of exact in-context
retrieval with fixed state~\citep{rnnsnottransformers} are special cases of this
picture.

Importance heuristics are best understood as distortion surrogates. No method can
evaluate $\dist$ directly, because $Y$ is unknown at compaction time. Each instead
picks a \emph{surrogate}, a cheap proxy for ``how much will dropping this hurt the
answer.'' Seen through Eq.~\eqref{eq:obj}, the field's apparently unrelated
heuristics are all surrogate distortion estimators: accumulated attention mass
(H2O, SnapKV;~\citealp{h2o,snapkv}), token self-information or perplexity
(LLMLingua;~\citealp{llmlingua}), an explicit mutual-information estimate
(QUITO-X;~\citealp{quitox}), a bound on the induced output error
(Ada-KV;~\citealp{adakv}), and an LLM's own judgment of salience (Mem0,
reflection;~\citealp{mem0,generativeagents}). Naming them as estimators of one
quantity is precisely what makes them comparable, and transferable, across
layers.

Three first-class properties separate methods that fail from those that do not. Two schemes can sit at the same budget
$\budget$ and the same average distortion yet behave completely differently on the
queries that matter. The difference is captured by three properties that
Eq.~\eqref{eq:obj} leaves implicit and that we promote to the center of the survey.
\textbf{(P-rev) Reversibility:} can content dropped by $\Cop$ be re-derived if a
later query needs it? Retrieval-backed and archival schemes can; eviction and
summarization cannot. \textbf{(P-q) Query-conditioning:} does $\Cop$ see the query
(or its distribution) when it decides what to keep? Offline gisting does not;
LongLLMLingua and Quest do. \textbf{(P-fid) Fidelity profile:} is the
representation lossless, near-lossless, uniformly lossy, or multi-fidelity (a small
high-fidelity tier plus a large lossy tier)?

The formalism yields four falsifiable predictions. The bound implies that violating
(P-rev)+(P-q) produces the same failure at every layer, with a
layer-specific, testable signature. \textbf{(1)~KV:} for attention-score eviction,
needle recall collapses once the per-token budget falls below the needle's share
of attention mass, and the collapse point tracks the entropy of the answer's
location, worsening in mid-context. \textbf{(2)~Prompt:} query-aware compression
shifts the accuracy-versus-ratio curve to the right of query-agnostic compression
by exactly the mutual-information gap. \textbf{(3)~Architectural:} a state of $s$
bits cannot answer a query requiring more than $s$ bits, so state-space and
linear-attention models exhibit a hard accuracy cliff on multi-key recall at a key
count proportional to state size, while attention degrades gracefully.
\textbf{(4)~Agent:} under repeated \emph{irreversible} summarization, end-task
error grows super-linearly in the number of compaction events, whereas a
reversible, retrieval-backed memory stays flat. We test (1) and (4) directly in
Section~\ref{sec:exp}.

\begin{figure}[t]
\centering
\begin{tikzpicture}[>=Latex,scale=0.95]
  \draw[->] (0,0) -- (5.5,0) node[below right,font=\scriptsize]{budget $\budget$ (bytes/token)};
  \draw[->] (0,0) -- (0,3.8) node[above,font=\scriptsize]{task utility};
  \draw[dashed,gray] (0,3.15) -- (5.3,3.15);
  \node[gray,anchor=east,font=\scriptsize] at (5.3,3.4){full-context utility};
  \draw[thick,blue] plot[smooth,tension=0.7] coordinates {(0.15,0.35)(0.7,1.55)(1.45,2.55)(2.5,2.98)(4.3,3.12)};
  \draw[thick,red,densely dashed] plot[smooth,tension=0.7] coordinates {(0.15,0.18)(1.15,0.7)(2.05,1.7)(3.2,2.62)(4.8,3.08)};
  \draw[dotted] (1.3,0) -- (1.3,3.15);
  \node[anchor=north,font=\scriptsize] at (1.3,0){$\Istar(Q)$};
  \draw[<->,thin] (1.55,1.3) -- (2.55,1.3);
  \node[font=\scriptsize,anchor=south,inner sep=1pt] at (2.05,1.3){$H(Q)$};
  \node[blue,font=\scriptsize,anchor=west] at (1.7,2.78){query-conditioned};
  \node[red,font=\scriptsize,anchor=west] at (2.85,2.0){query-agnostic};
  \node[gray,font=\scriptsize,align=center,anchor=south] at (0.66,0.05){forced\\[-1pt]error};
\end{tikzpicture}
\caption{The rate--distortion view of Section~\ref{sec:formalism}. Below the
task-conditioned information content $\Istar(Q)$, no scheme can avoid error
(Eq.~\ref{eq:bound}). A query-agnostic operator (red) reaches a given utility only
at a larger budget than a query-conditioned one (blue). The horizontal gap is the
query entropy $H(Q)$, which the bound charges for not knowing the query in advance.}
\label{fig:rd}
\end{figure}
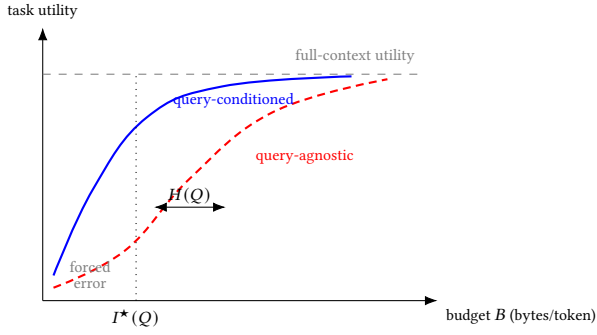

\section{A Seven-Axis Taxonomy}
\label{sec:taxonomy}

The formalism exposes the dimensions along which methods genuinely differ. We
pull these out as seven axes, chosen to be as close to orthogonal as the design
space allows. A method is a point in their product. As
Section~\ref{sec:formalism} predicts, a method's failure mode follows from the
combination of axis values, not from any single value read in isolation.

\begin{enumerate}\itemsep2pt
\item \textbf{Granularity of the compressed unit.} From finest to coarsest:
  bit-width (quantization), hidden dimension or rank (low-rank factorization),
  token / KV entry, page or block, layer or head, natural-language span, dense
  soft token, recurrent state, semantic item (a fact, note, or graph node), visual
  token, and inter-agent message.
\item \textbf{Lifecycle stage, where the operator acts.} Architecture and
  pretraining; prompt or prefill time, before the model reads the context;
  prefill-to-decode KV formation; decode-time dynamic selection; the serving
  runtime, across requests; within-task working-context curation; between-task
  consolidation; and offline corpus indexing.
\item \textbf{Lossiness and fidelity.} Lossless reuse, near-lossless
  approximation, uniformly lossy, or multi-fidelity (a small exact tier plus a
  large lossy tier); and, cross-cutting, reversible versus irreversible (P-rev).
\item \textbf{Query/task adaptivity.} Query-agnostic and offline-cacheable,
  query-conditioned and online, or task-aware via a learned reward (P-q); together
  with the importance \emph{signal} used (attention score, perplexity,
  mutual information, output-error bound, or LLM-judged salience), which the
  formalism identifies as competing distortion surrogates.
\item \textbf{Learnability.} Training-free heuristic, post-training adapter,
  trained-from-scratch architecture, RL-learned policy, or LLM-as-controller with
  no weight change.
\item \textbf{Mechanism.} Drop/evict, select/retrieve (keep all), merge/cluster,
  quantize/encode, factorize/low-rank, abstractive summarize/rewrite,
  encode-to-latent, recurrent write--forget, structure/graph build, or
  internalize-to-weights.
\item \textbf{Storage substrate.} On-GPU KV, a host/SSD/remote tier, model
  parameters, an external text or vector store, a knowledge graph, in-context dense
  vectors, or a hybrid parametric/non-parametric store.
\end{enumerate}

Table~\ref{tab:layers} gives the characteristic axis values for each layer we
survey. The differences sit in three columns: granularity, lifecycle, and
adaptivity. These are the three dimensions the lens flags as decisive, and the
remaining axes vary far less across layers. Appendix~\ref{app:master} carries
the full per-method classification of roughly seventy methods.

\begin{table*}[t]
\centering
\small
\setlength{\tabcolsep}{4.5pt}
\renewcommand{\arraystretch}{1.25}
\begin{tabular}{l p{2.2cm} p{2.5cm} p{2.3cm} p{2.4cm} p{2.6cm}}
\toprule
\textbf{Layer} & \textbf{Granularity} & \textbf{Lifecycle stage} & \textbf{Typical fidelity} & \textbf{Adaptivity} & \textbf{Mechanism / substrate} \\
\midrule
KV-cache (\S\ref{sec:kv}) & bit, rank, token, head & prefill-to-decode, decode-time & near-lossless, usually irreversible & mostly query-agnostic & evict / quantize / low-rank / merge; on-GPU KV \\
Prompt/context (\S\ref{sec:prompt}) & NL span, soft token & pre-model (prefill) & lossy (hard) or opaque (soft) & agnostic or query-aware & prune / summarize / encode; in-context \\
Architectural (\S\ref{sec:arch}) & recurrent state & pretraining & uniformly lossy, fixed size & learned & recurrent write--forget; parameters \\
Agent/semantic (\S\ref{sec:agent}) & fact, note, summary & within- and between-task & lossy, partly reversible & LLM- or RL-driven & summarize / structure; external store \\
Cross-cutting (\S\ref{sec:sparse},\,\S\ref{sec:multi}) & block, visual token, message & pretraining; pre-model; cross-agent & learned or highly lossy & learned / similarity-based & learned sparsity / merge; mixed substrate \\
\bottomrule
\end{tabular}
\caption{The layers we survey and their characteristic values on the seven axes
(Section~\ref{sec:taxonomy}). All layers share the axes; what separates them is
granularity, lifecycle stage, and whether the retain decision is
query-conditioned, the three dimensions the unifying lens singles out as
decisive.}
\label{tab:layers}
\end{table*}

\begin{figure*}[t]
\centering
\begin{tikzpicture}[>=Latex,
  st/.style={draw,rounded corners,fill=teal!10,minimum height=8mm,text width=2.05cm,align=center,font=\small\bfseries},
  ex/.style={text width=2.05cm,align=center,font=\scriptsize}]
\node[st](a){Pretraining};
\node[st,right=5mm of a](b){Prompt / prefill};
\node[st,right=5mm of b](c){Decode};
\node[st,right=5mm of c](d){Serving};
\node[st,right=5mm of d](e){Within-task};
\node[st,right=5mm of e](f){Cross-task};
\foreach \x/\y in {a/b,b/c,c/d,d/e,e/f}{\draw[->,thick,gray] (\x.east)--(\y.west);}
\node[ex,below=2mm of a]{Mamba, MLA, RMT, NSA};
\node[ex,below=2mm of b]{LLMLingua, gisting, SnapKV};
\node[ex,below=2mm of c]{H2O, Quest, KIVI, TOVA};
\node[ex,below=2mm of d]{PagedAttention, InfiniGen};
\node[ex,below=2mm of e]{ACON, MEM1, Context-Folding};
\node[ex,below=2mm of f]{Mem0, RAPTOR, sleep-time};
\end{tikzpicture}
\caption{Where compaction acts across the model and agent lifecycle, the lifecycle
axis of Section~\ref{sec:taxonomy}. The same retain-versus-discard decision recurs
at each stage, on a different substrate and timescale; representative methods appear
beneath each.}
\label{fig:lifecycle}
\end{figure*}
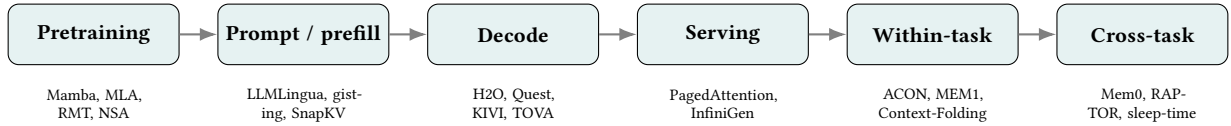
\section{KV-Cache-Level Compaction}
\label{sec:kv}

The KV cache attracts more compaction work than any other layer. At long context
and large batch it is the cache, not the weights, that dominates GPU memory and
decode bandwidth, since every layer rereads every cached key and value at every
step. The history $\Hist$ is the cache itself: per layer, per head, a key and
value vector for each past token. The operator $\Cop$ shrinks $\rate(\Zrep)$
measured in GPU bytes, and $\Uop$ is ordinary attention over whatever survives.
The dozens of methods below share one device, the \emph{surrogate} they use for
distortion. Nearly all of them estimate ``how much will dropping or coarsening
this entry hurt the answer'' from the attention score, which
Eq.~\eqref{eq:obj} casts as a cheap distortion proxy, biased in the ways the
bound predicts. We group methods by which axis of the cache they compress: the
token (sequence) axis, the query-time read pattern, the bit-width, the hidden
dimension, and the layer axis.

\subsection{Token Eviction: Dropping Entries Irreversibly}

Eviction permanently deletes a budgeted subset of tokens' KV, exploiting the
empirical sparsity of attention, in which a small set of tokens carries most of the mass.
Every method here is irreversible (violates (P-rev)) and, with one exception,
decides what to keep before the decode query is known (violates (P-q)); they are
therefore the canonical instantiation of prediction~(1).

The cheapest evictors ignore content entirely and rely on positional or structural
policies. StreamingLLM~\cite{streamingllm} keeps the first few
``attention-sink'' tokens plus a sliding window of recent tokens and discards the
middle, giving constant memory and stable infinite streaming for free; its cost is
total amnesia about evicted middle context, so it fails any task whose evidence
lies outside the window. The sink it exploits is itself now well understood as a
structural property of softmax attention~\cite{attnsinkemerges,sinknecessity}.
FastGen~\cite{fastgen} adds content awareness without scoring: a one-shot
profiling pass classifies each head as local, special-token, punctuation, or
broad, and applies a per-head policy matched to that structure, recovering most
attention at roughly half the memory but yielding only modest gains over
fixed-budget scorers.

Most evictors instead score each token by its
accumulated or recent attention and keep the top-$\budget$. H2O~\cite{h2o} casts
eviction as greedy submodular maximization of accumulated attention, retaining a
balance of recent and ``heavy-hitter'' tokens; Scissorhands~\cite{scissorhands}
leans on a \emph{persistence of importance} hypothesis, that tokens pivotal once stay
pivotal, to fix a budget of historically high-attention tokens. Both read
importance only from past steps. They therefore over-retain early, long-lived
tokens and miss saliency that shifts with the (unseen) query. TOVA~\cite{tova} reframes the
decoder as a bounded multi-state RNN and drops the single lowest-current-attention
token per step, removing recency assumptions but making a transient dip in
attention permanently fatal. Keyformer~\cite{keyformer} corrects the distribution
shift that dropping induces with a Gumbel-softmax-regularized score, and
SnapKV~\cite{snapkv} introduces the influential \emph{observation window}: it
pools attention from the last prompt tokens to select, per head, the most-attended
prompt tokens before generation. SnapKV works precisely when the query is already
visible at prompt end, a partial, implicit form of query-conditioning, and
degrades when it is not, exactly the gap of size $H(Q)$ that the bound assigns to
not knowing the query.

A related question is not what to keep but
where to spend a fixed total budget. PyramidInfer~\cite{pyramidinfer} and
PyramidKV~\cite{pyramidkv} allocate layer-wise on a ``pyramidal information
funneling'' observation, since attention is broad in lower layers and focused in
higher ones. They give large budgets to shallow layers and small ones to deep layers,
then apply SnapKV-style selection within each. Ada-KV~\cite{adakv} reallocates
\emph{across heads}: a layer-global top-$k$ assigns each head a budget proportional
to its share of the global selection, and it is the rare method that ties its
surrogate to theory, deriving the allocation from a bound on post-eviction
attention-output error rather than raw attention mass. These are
allocation layers stacked on existing scorers; they improve budget utilization but
inherit the underlying query-agnostic bias, and the layer profile remains a
heuristic.

\subsection{Dynamic Sparse Retrieval: Keeping Everything, Reading Selectively}

A distinct strategy keeps the full cache in storage and instead makes the
\emph{read} sparse and query-conditioned at each step, restoring (P-rev) and
(P-q) at the price of capacity savings. Quest~\cite{quest} stores all KV in pages
and, per decode step, estimates each page's criticality from min/max key bounds
against the current query, loading only the top-$K$ critical pages; because no
token is ever discarded and selection adapts to the query, it is near-lossless
where eviction collapses. Landmark Attention~\cite{landmark} trains representative
``landmark'' tokens that gate random access to blocks of context, achieving the
same effect through a learned index. These methods cut bandwidth, not capacity:
they convert ``what to forget'' into ``what to fetch,'' which is why the bound's
$\Istar(Q)$ penalty does not bind on them as it does on evictors.

\subsection{Quantization: Compressing the Bit-Width}

Orthogonal to which entries survive is how many bits each costs. Quantization is
near-lossless in the bound's terms, since it preserves (P-rev) as graceful fidelity
loss rather than deletion, and is the most mature sub-field. KIVI~\cite{kivi} is
the reference recipe: tuning-free asymmetric 2-bit quantization that quantizes
keys \emph{per channel} and values \emph{per token} to match their distinct
outlier structures, keeping a small full-precision residual window.
KVQuant~\cite{kvquant} pushes to 2--3 bits with per-channel, pre-RoPE,
sensitivity-weighted non-uniform quantization plus dense-and-sparse outlier
isolation, reaching million-token contexts on one GPU. Coupled
Quantization~\cite{coupledquant} approaches one bit per channel by jointly
encoding groups of statistically inter-dependent channels with a shared code.
Atom~\cite{atom} takes the systems view, quantizing weights, activations, and KV
to 4 bits end-to-end with co-designed INT4 kernels so the memory saving becomes a
throughput saving. The recurring obstacle, flagged across this family, is RoPE:
rotary embeddings smear per-channel statistics, forcing the pre-RoPE quantization
and decoupled handling that complicate kernels.

\subsection{Low-Rank and Hidden-Dimension Compression}

Compression can also target the per-token feature dimension. Palu~\cite{palu}
post-training low-rank-decomposes the K/V projections so only a smaller latent is
cached and full keys/values are reconstructed on the fly, with a rank-search
procedure and quantization compatibility, roughly halving the cache on existing
checkpoints. GEAR~\cite{gear} is explicitly compositional: ultra-low-bit
quantization of the bulk, a low-rank matrix approximating the residual
quantization error, and a sparse matrix for outliers, recovering much of the
accuracy lost at 2 bits. Architecturally, Multi-head Latent Attention (MLA) in
DeepSeek-V2~\cite{mla} \emph{trains in} low-rank joint KV compression, caching one
small latent up-projected at inference; it matches or beats MHA quality with a far
smaller cache and has become the de facto standard for new frontier models, at the
cost of pretraining from scratch and decoupled-RoPE handling. Bridging post-hoc
and architectural, TransMLA~\cite{transmla} seeks cheap conversion of existing GQA
checkpoints into latent attention, attacking the retrofittability gap.

\subsection{Cross-Layer Sharing and Token Merging}

Redundancy also runs along the layer dimension and within surviving tokens.
xKV~\cite{xkv} performs post-training cross-layer SVD, consolidating a group of
layers' caches into a shared low-rank subspace by exploiting alignment of their
dominant singular vectors, reaching up to $8\times$ reduction. Architecturally,
Cross-Layer Attention (CLA)~\cite{cla} has only some layers compute K/V and the
rest reuse earlier layers' activations (orthogonal to MQA/GQA), and YOCO (You Only
Cache Once)~\cite{yoco} caches KV exactly once via a decoder-decoder design in
which a stacked cross-decoder reuses one global cache, saving roughly an $L$-fold
factor. MiniCache~\cite{minicache} is the training-free counterpart, merging
highly similar KV states between adjacent middle-to-deep layers by interpolating
direction while preserving magnitude. A complementary idea merges rather than
drops tokens: CaM~\cite{cam} adaptively fuses to-be-evicted KV into surviving
entries weighted by the discarded tokens' attention, recovering much of eviction's
output perturbation, and KVMerger clusters similar consecutive tokens into a
pivotal token via Gaussian-kernel weights. Merging partially restores (P-rev) by
averaging rather than deleting, but blurs distinct long-range information at small
budgets.

\begin{table}[t]
\centering
\small
\setlength{\tabcolsep}{3.2pt}
\renewcommand{\arraystretch}{1.15}
\begin{tabular}{l l l ccc}
\toprule
\textbf{Method} & \textbf{Mechanism} & \textbf{Gran.} & \textbf{Q-c.} & \textbf{Rev.} & \textbf{Reduct.} \\
\midrule
StreamingLLM~\cite{streamingllm} & positional evict & token  & \none & \none & const.\ mem \\
H2O~\cite{h2o}                   & attn-score evict & token  & \none & \none & $\sim$5$\times$ \\
SnapKV~\cite{snapkv}             & obs-window evict & head   & \half & \none & $\sim$8$\times$ \\
PyramidKV~\cite{pyramidkv}       & layer-wise evict & layer  & \half & \none & $\sim$8$\times$ \\
Ada-KV~\cite{adakv}              & head-wise budget & head   & \half & \none & varies \\
Quest~\cite{quest}              & sparse retrieval & page   & \full & \full & read-only \\
KIVI~\cite{kivi}                & 2-bit quant      & ch./tok & \none & \full & $\sim$2.6$\times$ \\
KVQuant~\cite{kvquant}          & non-unif.\ quant & channel & \none & \full & $\sim$5$\times$ \\
Palu~\cite{palu}                & low-rank proj    & hidden & \none & \full & $\sim$2$\times$ \\
MLA~\cite{mla}                  & latent attn      & arch.\  & \none & \full & $\gg$MHA \\
CaM~\cite{cam}                  & token merge      & token  & \none & \half & add-on \\
\bottomrule
\end{tabular}
\caption{Representative KV-cache compaction methods by mechanism axis.
Query-conditioning (Q-c.) and reversibility (Rev.): \full~yes, \half~partial,
\none~no. Reductions are reported KV savings from the cited works and are not
directly comparable across setups. Eviction methods are query-agnostic and
irreversible, instantiating prediction~(1); retrieval and quantization preserve
reversibility.}
\label{tab:kv}
\end{table}

\subsection{Cross-Cutting Tensions}

Three problems run through Table~\ref{tab:kv}. The attention-score surrogate is
systematically biased: accumulated scores over-retain early and high-frequency
tokens and miss collectively-important marginal ones, which is why output-error-aware
scoring (Ada-KV) and merging (CaM) have started to appear as corrections. RoPE
interferes everywhere, confounding per-channel quantization and low-rank latents
alike; that only some ``retrieval heads'' carry long-range dependence has
motivated head-specific treatment (RazorAttention~\cite{razorattention},
ShadowKV~\cite{shadowkv}). The third problem matters most for our thesis. Methods
that look near-lossless on summarization and Needle-in-a-Haystack fall off sharply
on retrieval, multi-hop, long-generation, and reasoning/long-CoT workloads, which
are exactly the high-$\Istar(Q)$ tasks the bound says will resist compression; this
gap has prompted decode-aware variants such as SCOPE~\cite{scope}. Composition is
the open frontier. Quantization, low-rank, and eviction attack orthogonal axes,
yet whether their errors compound or cancel, and which combinations help, remains
largely uncharted, and it is where a rate--distortion account should be doing the
design work.

\section{Context and Prompt Compaction}
\label{sec:prompt}

Prompt compaction works on the \emph{input itself}, shrinking $\Hist$ before the
target model attends to it. It chooses $\Zrep$ in the currency of input tokens or
of soft embeddings that occupy token slots. KV eviction prunes inside a fixed
forward pass; prompt compaction instead commits to a representation the target
LLM must re-encode, so its design problem is Eq.~\eqref{eq:obj} taken literally:
maximize $\I(\Zrep; Y \mid Q)$ at a token budget $\budget$. A dedicated survey
already catalogs this literature by method family~\citep{surveypromptcomp}. Our
reading runs it through the IB objective of Section~\ref{sec:formalism} and places
it against the other layers, which surfaces the axis the field's own taxonomy
underplays: whether $\Zrep$ stays human-readable text or becomes opaque dense
vectors. We split the section along that line.

\subsection{Hard compaction: extractive and abstractive text}
\label{sec:prompt-hard}

Hard methods keep $\Zrep$ in natural language, deleting low-value tokens or
rewriting the context into fewer words. The output is text, so the target model
stays untouched, the artifact ports across models, and a human can read it. That
last property, inspectability, is reversibility's weaker cousin. The shared move
is to approximate the distortion in Eq.~\eqref{eq:obj} with a
\emph{token-importance surrogate} from a small auxiliary model, then keep the
high-importance tokens.

The first generation
estimates importance from a language model's own uncertainty, using self-information
and perplexity surrogates. Selective Context
scores each lexical unit by its self-information $-\log p(x_i\mid x_{<i})$ under a
base LM and drops the least surprising tokens, phrases, or
sentences~\citep{selectivecontext}; LLMLingua makes this iterative and
budget-controlled, using a small LM's token perplexity in a coarse-to-fine pass
that protects entities and numbers while pruning predictable
filler~\citep{llmlingua}. Self-information and perplexity are exactly the
distortion surrogates named in Section~\ref{sec:formalism}: a token the model can
already predict carries few bits of $\I(\Zrep;\Hist)$ and is cheap to drop. Their
weakness is that they score importance against the corpus prior rather than the
query, a query-agnostic operator that, by consequence~(iii) of
Eq.~\eqref{eq:bound}, must spread $\budget$ across the whole query distribution.
LongLLMLingua closes this gap by conditioning the surrogate on $Q$: it ranks
content by question-aware contrastive perplexity, reallocates per-document budget,
and reorders documents to fight lost-in-the-middle~\citep{longllmlingua}. This is
the cleanest empirical instance of prediction~(2): moving from corpus-prior to
query-conditioned scoring shifts the accuracy-versus-ratio curve to the right by
the mutual-information gap $H(Q)$, which is why LongLLMLingua can match or exceed
the uncompressed prompt at ratios where LLMLingua already degrades.

Perplexity scoring is slow
and tokenizer-mismatched to the target, so the second generation replaces it with a
learned classifier, distilled or RL-trained. LLMLingua-2 reframes compaction as bidirectional
preserve/discard token classification, training a small encoder on a GPT-4 distilled
extractive dataset; it is faster and more faithful than perplexity pruning and is
now the de facto task-agnostic baseline~\citep{llmlingua2}, but, supervised by a
fixed teacher, it inherits a query-agnostic notion of importance. TACO-RL keeps the
classifier but trains it with reinforcement learning against a task-specific reward
(output divergence under a ratio constraint), turning a static surrogate into one
optimized for the actual downstream loss in Eq.~\eqref{eq:obj}~\citep{tacorl}. The
trajectory from Selective Context through LLMLingua-2 to TACO-RL is the field walking
its surrogate from a corpus-prior proxy toward the true task distortion, paying for
each step with a more specialized, less transferable compressor.

QUITO-X makes the
objective literal with an explicit information-bottleneck criterion. Rather than a
perplexity proxy, it uses a cross-attention module to score each context token by
its mutual information with the answer conditioned on the question, pruning to
maximize the retained query-relevant $\I(\Zrep; Y \mid Q)$, which instantiates the
IB form of Eq.~\eqref{eq:obj} directly~\citep{quitox,tishbyib}. It is the clearest evidence that the
heterogeneous ``importance'' heuristics of this layer are all estimators of one
quantity; the others differ only in how crudely they approximate the same MI.

A parallel line compacts retrieved evidence,
where budget pressure is acute and the query is always present. RECOMP pairs an
extractive selector with an abstractive summarizer trained for query relevance,
emitting an empty summary when retrieval is unhelpful~\citep{recomp}; FILCO trains a
sentence filter supervised by lexical and conditional cross-mutual-information
signals~\citep{filco}; Provence casts pruning as sequence labeling over the passage
with a reused reranking head, making it near-zero-cost and self-calibrating in how
many sentences to keep~\citep{provence}; and CompAct condenses documents iteratively
and actively for multi-hop QA, reaching extreme ratios at the cost of per-query
generation and summarizer hallucination~\citep{compact}. These span the granularity
spectrum, from sentence-level filtering (FILCO, Provence) through single-pass
abstractive summarization (RECOMP) to iterative active abstraction (CompAct),
but are all query-conditioned by construction, which is why they sit far right on the
accuracy-ratio curve. Their limit is faithfulness: abstractive variants give no
guarantee that dropped facts were truly irrelevant, an irreversible loss no later
query can recover (violating P-rev).

\subsection{Soft compaction: learned latent representations}
\label{sec:prompt-soft}

Soft methods let $\Zrep$ leave the token vocabulary: they train the context into a
handful of dense vectors (gist tokens, memory slots, or compressed KV states) that
occupy token slots but are continuous. Freed from the discreteness of text, they
reach far higher ratios, but $\Zrep$ becomes opaque (no human can audit it) and
typically query-agnostic and model-coupled.

The founding idea routes a prompt's
information through a few appended slots, an autoencoder from prompt to gist. Gisting trains the LLM during instruction
tuning with a restricted attention mask that forces context to flow through ``gist''
tokens, yielding reusable, cacheable activations~\citep{gisting}; AutoCompressor
recursively summarizes segments into accumulating summary vectors for very long
inputs~\citep{autocompressor}; ICAE uses a LoRA-adapted copy of the LLM as an
encoder that compresses context into memory slots decoded by the frozen
model~\citep{icae}; and 500xCompressor pushes to as few as one token by encoding
into compressed KV (not output) representations, finding KV preserves more than
embeddings at extreme ratios~\citep{compressor500x}. These are autoencoders for
context: their reconstruction objective biases $\Zrep$ toward retaining \emph{all}
of $\Hist$ rather than the answer-relevant bits, which is why, by the
task-dependence of Eq.~\eqref{eq:bound}, they are near-lossless on redundant prose
yet shed exact-recall capacity sharply as the budget falls.

Some methods compress the per-layer cache rather
than the input, distilling KV into latents. Activation Beacon interleaves beacon tokens that progressively
compress each layer's key/value activations, extending context to hundreds of
thousands of tokens at fixed quality~\citep{activationbeacon}; KV-Distill distills a
long KV cache into a short, question-independent one by treating the two caches as a
student-teacher pair under a KL objective~\citep{kvdistill}. Being
query-independent buys offline caching but forfeits the per-query budget advantage
of consequence~(iii); these methods pay the $H(Q)$ price by construction.

At the extreme, xRAG
reinterprets a precomputed document retrieval embedding as a single modality token
projected into the LLM's space, with retriever and model frozen~\citep{xrag}: one
vector, capacity hard-capped at the embedding. Context Distillation goes the
opposite way, internalizing context into the weights for zero inference tokens, at
the price of per-prompt retraining and no swappable artifact~\citep{contextdistill}.
Cartridges generalizes this via offline per-corpus self-study, amortizing a long
fixed corpus into a small trained KV ``cartridge'' reused across
queries~\citep{cartridges}, a query-agnostic compaction whose cost is hidden from
the serving path, akin to sleep-time compute~\citep{sleeptime}.

Hard against soft trades interpretability and portability for compression ratio.
Soft tokens buy the higher ratio with opacity and query-agnosticism, and they run
a real risk: a single opaque vector can silently drop the bits a future query
needs. The most rigorous study of
gist compression confirms this: it is near-lossless on RAG and long-document QA but
exhibits systematic failures (``lost by the boundary,'' ``lost if surprise,'' and
``lost along the way'') on synthetic recall, and its fine-grained mitigations
narrow but do not close the gap to full attention~\citep{silverbullet}. That work also
shows a plain average-pooling baseline often rivals trained
gisting on long inputs: when $\Hist$ is information-dense, the learned compressor
extracts little beyond what naive pooling captures, indicating the paradigm is not
yet a drop-in for attention. Read through Eq.~\eqref{eq:bound}, these failures are
the expected signature of a high-$\Istar(Q)$ task met by a query-agnostic
operator, the same wall the other three layers hit, here wearing soft-token
clothing.

\section{Architectural and Bounded-State Compaction}
\label{sec:arch}

Sections~\ref{sec:kv} and~\ref{sec:prompt} compact a representation the model was
never trained to produce. They evict, quantize, or rewrite a cache the architecture
builds at full size. The methods here instead bake compaction into the model and
train it end to end, so the bounded state \emph{is} the compact representation
$\Zrep$ and the operators $\Cop_\theta$, $\Uop$ are learned jointly with the
weights. They sit at the learned, trained-from-scratch end of the learnability
axis, and they push the rate--distortion tension to its limit. When the state is a
fixed $s$-bit object updated online, the bound of Eq.~\eqref{eq:bound} bites
directly: $\budget = s$ is fixed once, before any query is seen. This is
prediction~(3) of Section~\ref{sec:formalism}. A state of $s$ bits cannot answer a
query whose answer requires more than $s$ bits, and what separates these methods is
how each tries to spend $s$ well.

The earliest architectural move keeps no compressed state at all; it caches.
Transformer-XL caches the hidden activations of the previous fixed-length segment
and reuses them as gradient-stopped keys and values for the current one, giving
segment-level recurrence past the training window~\citep{transformerxl}. The cache
is lossless but offers no compression: rate grows linearly with the number of
cached segments, so the cache must be capped and older context dropped, a
hard-edged, query-agnostic eviction (violating (P-rev) and (P-q)). The Compressive
Transformer adds a second, multi-fidelity tier (P-fid). Activations evicted from
short-term memory are compressed at a fixed rate, via a strided convolution trained
with an attention-reconstruction loss, into a smaller long-term
store~\citep{compressivetransformer}. The Recurrent Memory Transformer (RMT)
replaces caching with a true bottleneck. A small set of learned \emph{memory}
tokens carries information from one segment to the next, their outputs feeding back
as the next segment's memory inputs, with no change to the core block~\citep{rmt}.
The number of memory tokens fixes $s$ exactly, which makes RMT the cleanest
instance of a learned, fixed-rate $\Cop_\theta$ and the cleanest test of where the
bound bites.

Another strand keeps the state lossy but makes the loss reversible (P-rev),
retrieving from an external store instead of compressing into a bottleneck.
Memorizing Transformers augment one attention layer with an approximate
$k$-nearest-neighbor lookup into a large, non-differentiable memory of past
key--value pairs, attending to retrieved history without backpropagating through
it~\citep{memorizingtransformers}. RETRO instead retrieves text chunks from a
trillion-token database and fuses them through chunked cross-attention, decoupling
knowledge from parameters~\citep{retro}. Landmark Attention inserts a learned
summary (``landmark'') token per block and trains attention to first select
relevant blocks by their landmarks and then attend within them, giving random
access through the attention mechanism itself~\citep{landmark}. The shared caveat,
in the formalism's terms, is that these methods sparsify retrieval but not storage.
The full KV still lives somewhere, so the true $\rate(\Zrep)$ is unchanged and the
``compression'' is of compute, not memory.

A third family makes the long-term store genuinely bounded. Infini-attention writes
keys and values evicted from a local window into a fixed-size associative memory
matrix by a linear-attention (delta-rule) update and reads them back with linear
attention, fusing local softmax attention and a constant-size long-term memory in
one block~\citep{infiniattention}. Memory is $O(1)$, but the read is the lossy
linear-attention reconstruction that softmax avoids, and the write lacks a strong
decay/forget gate. The method proved hard to reproduce, which the formalism reads
as a missing content-aware write policy: it spends $s$ without a surrogate for what
to overwrite. Activation Beacon makes the same fixed-size move as a trainable
plug-in on a frozen base, condensing each chunk's keys and values into a few
``beacon'' activations via compression-based
autoregression~\citep{activationbeacon}.

The dominant 2023--2025 strand treats the recurrent state itself as the compressor.
Mamba replaces attention with an input-dependent (selective) state-space recurrence
whose fixed-size hidden state summarizes all history, with a hardware-aware parallel
scan for linear-time training and constant-state inference~\citep{mamba}. RWKV
reaches the same constant-state regime as a linear-attention ``RNN-Transformer''
with a time-decayed key--value state~\citep{rwkv}. Both make the rate--distortion
trade explicit: the constant state is a lossy summary, so exact long-range copy and
in-context recall lag softmax attention, which is prediction~(3) made concrete. A
delta-rule lineage refines how the state is written. DeltaNet replaces additive
accumulation with a delta-rule (error-correcting) update that overwrites the value
bound to a key, and parallelizes it across the sequence~\citep{deltanet}; Gated
DeltaNet adds a data-dependent forget gate atop the delta rule, combining
principled erasure with decay~\citep{gateddeltanet}; GLA supplies hardware-efficient
gated linear attention~\citep{gla}; and xLSTM revives LSTM gating with exponential
gates and a matrix memory~\citep{xlstm}. All four learn a write/forget policy, a
surrogate for which bits of $s$ to overwrite, the very policy Infini-attention
lacked.

Constant state buys efficiency at the cost of exact recall, and deployed systems
respond by keeping a thin layer of softmax attention for precise short-range access
while delegating the long tail to cheap recurrent state, a multi-fidelity (P-fid)
split across layers. Jamba interleaves Transformer and Mamba blocks with a
mixture-of-experts feed-forward~\citep{jamba}; Samba alternates Mamba layers with
sliding-window attention~\citep{samba}; Griffin mixes gated linear recurrences with
local attention~\citep{griffin}. Each spends a small precise budget where queries
need random access and a large lossy budget where they need only gist, an
architectural realization of the prediction that query-relevant bits should be held
at high fidelity and the rest at low.

A further strand makes the write policy itself adaptive at inference. Titans makes
the recurrent memory a deep module whose weights are updated at test time by
gradient descent on a ``surprise'' (memory-prediction-error) signal with momentum
and adaptive forgetting, scaling to multi-million-token contexts~\citep{titans}.
This is a learned $\Cop_\theta$ that re-optimizes its own distortion surrogate
online. Larimar couples a decoder with a Kanerva-style distributed associative
memory, enabling one-shot, training-free writes, edits, and selective forgetting at
inference, at the cost of a lossy associative read bounded by the memory
dimension~\citep{larimar}. MemoryLLM embeds a fixed-size pool of latent memory
tokens in every layer that self-updates as new text arrives, with controlled
forgetting~\citep{memoryllm}. Its pool is again a fixed $\budget$, and retention
degrades once history exceeds it, the bound observed empirically.

Two methods convert an existing model into a bounded-state one without retraining
from scratch. Dynamic Memory Compression (DMC) learns, per head and via continued
pretraining, whether to append a new key--value pair or accumulate it into the
previous slot, retrofitting an adaptive, content-aware compression rate onto a
standard cache~\citep{dmc}. TransMLA converts a pretrained grouped-query-attention
model into a multi-head latent attention model~\citep{transmla,mla}, replacing the
GQA cache with the low-rank latent state of MLA. This swaps the currency of
$\rate(\Zrep)$ after the fact, bridging this section's trained-from-scratch end to
the inference-time factorizations of Section~\ref{sec:kv}.

\paragraph{Constant memory versus exact recall.} Every mechanism here runs into the
same constraint. A fixed $s$-bit state cannot answer queries whose answers require
more than $s$ bits, and no write policy escapes the data-processing inequality
behind Eq.~\eqref{eq:bound}. The surrogate that decides which bits to keep can
improve; the ceiling cannot. Nor is the bound merely asymptotic. RNNs with bounded
state are provably weaker than transformers at in-context
retrieval~\citep{rnnsnottransformers}, and RMT, precisely because its memory size
is an explicit knob, finds needles in an 11M-token haystack that full-attention
models miss, by spending its small budget on the relevant span and discarding the
rest~\citep{rmtneedles,babilong}. That result is the bound made positive: when
$\Istar(Q)$ is small and the operator is query-aligned, a tiny $s$ suffices; when
the query distribution is broad or answers carry many bits, no fixed $s$ does. So
the architectural strand chooses, at training time, where on the rate--distortion
frontier to sit, rather than evading the frontier altogether.

\section{Agent and Semantic-Memory Compaction}
\label{sec:agent}

This is the longest-timescale layer, and the one where the formalism's abstractions
turn into concrete operations. The history $\Hist$ here is the accumulated trace of
an agent's life: tool calls, observations, dialogue turns, and prior reflections,
often spread across many sessions, not a KV cache or a token window. The compact
representation $\Zrep$ is an external, mostly natural-language store of summaries,
extracted facts, structured notes, or a knowledge graph, and the usage operator
$\Uop$ is a retrieval step that pages a subset of $\Zrep$ back into the window
before generation. Two features set the layer apart. The budget $\budget$ is
enforced not by hardware but by the context window of $\Uop$: an unbounded store is
useless if the agent can only read a few thousand tokens of it, so the bottleneck is
retrieval, not storage. And the distortion surrogate is almost always
\emph{LLM-judged salience}. An LLM scores, summarizes, or decides what to keep, so
the compaction operator is itself a fallible model-in-the-loop rather than a cheap
statistic. This is expensive in two ways: every compaction event is an LLM call, so
the savings a smaller context buys are partly repaid in inference overhead, and any
error the judge commits is written into $\Zrep$ for good. The decisive axis here is
reversibility (P-rev). Schemes that keep the raw trace archived and retrievable are
reversible; schemes that overwrite it with a lossy summary are not, and the
irreversible ones drive prediction~(4).

The first family compacts the running trajectory of a single task as it approaches
the window limit. MemGPT/Letta casts this as operating-system paging: the LLM treats
its window as ``main memory'' and an external vector store as ``disk,'' and issues
function calls to evict, summarize, and page content back in~\citep{memgpt}. The OS
metaphor is exactly the rate--budget trade of Eq.~\eqref{eq:obj}, with $\budget$ the
window and the store unbounded but slow, and its weakness is the one the formalism
names: paging is governed by the model's own summaries, so eviction is irreversible
whenever the archival copy is itself a summary rather than the raw turn. ACON keeps
the operator training-free but makes the surrogate explicit. It compresses
observations and history under natural-language ``compression guidelines'' that are
iteratively refined from paired trajectories where the full context succeeds but the
compressed context fails, a failure-driven estimate of which dropped bits raised
distortion~\citep{acon}. The 2025--2026 frontier replaces hand-tuned operators with
RL-learned compaction policies that optimize a proxy for $\dist(\theta)$ directly.
Context-Folding branches into a sub-trajectory for a subtask and, on return, folds
away the intermediate steps, keeping only an outcome summary, trained end-to-end
with the FoldGRPO objective~\citep{contextfolding}; Memory-as-Action frames in-place
edits (prune-and-write deletions and insertions) as policy actions learned jointly
with task-solving~\citep{memact}; MEM1 carries a single constant-size internal state
that it rewrites every turn by merging prior memory with the new observation and
discarding redundancy~\citep{memone}; and ReSum periodically invokes a summarization
tool to collapse a web-search history into a compact reasoning state, with ReSum-GRPO
training the agent to reason fluently from summaries~\citep{resum}. IterResearch
rebuilds a streamlined workspace each round from only the essential prior outputs,
treating an evolving central report as the agent's Markovian
memory~\citep{iterresearch}. These methods report 3--10$\times$ active-context
reduction at matched accuracy. They also share a structural risk: all fold, prune,
and rewrite irreversibly, so a detail the policy judged unimportant is gone, and the
RL signal optimizes average distortion, not the tail-query distortion that the bound
shows is where low-budget memory fails.

The second family compacts across task boundaries, promoting raw episodes into
reusable knowledge. Its canonical pipeline is reflection. Generative Agents keep a
timestamped memory stream scored by recency, importance, and relevance, and
periodically synthesize salient memories into higher-level reflections that are
themselves stored and retrieved~\citep{generativeagents}; the importance score is
the prototypical LLM-judged salience surrogate. Reflexion specializes this to
failure, writing a verbal self-critique after a failed attempt and prepending it on
the next trial~\citep{reflexion}, and ExpeL distills cross-task experience by
contrasting successful and failed trajectories into natural-language
insights~\citep{expel}. Being gradient-free makes these methods cheap and general,
but it also exposes the layer's signature pathology: a wrong reflection or
over-generalized rule is stored and then biases every future retrieval. This
\emph{self-reinforcing error} is the agent-level analogue of an evicted needle, with
a twist: the agent does not just lose a good memory, it reconstructs a bad one on
demand. A second sub-family treats consolidation as structured editing. Mem0 runs
an extract-then-update pipeline in which an LLM pulls salient facts and then issues
ADD/UPDATE/DELETE/NOOP operations to keep a compact, consistent store~\citep{mem0};
A-MEM builds Zettelkasten-style atomic notes that are dynamically linked to and
evolve related notes into a self-organizing network~\citep{amem}; MemoryBank adds
explicit forgetting, decaying memories by an Ebbinghaus curve so that significance
and recency, not capacity alone, govern what survives~\citep{memorybank}; and MIRIX
routes information into six typed modules (core, episodic, semantic, procedural,
resource, knowledge-vault), each with its own fields and access
policy~\citep{mirix}. The UPDATE/DELETE operator is the cleanest realization of an
irreversible $\Cop$: a mis-merge or wrongful deletion cannot be re-derived, so its
error is permanent in exactly the sense (P-rev) isolates. Sleep-time compute reframes
the economics, moving consolidation offline so that the LLM ``thinks'' about a
context before any query arrives and caches useful inferences, amortizing the
per-compaction LLM-call cost across future queries~\citep{sleeptime}. The catch is
that it spends budget speculatively on bits that may never be queried, the dual of
the query-agnostic penalty $H(Q)$ in Eq.~\eqref{eq:bound}.

A third family makes the store itself structured, so that $\Uop$ retrieves
abstractions rather than raw chunks and compaction moves into an offline indexing
stage. RAPTOR recursively clusters and summarizes chunks into a bottom-up tree,
letting retrieval draw from multiple abstraction levels, a multi-fidelity (P-fid)
store with lossy summaries above lossless leaves~\citep{raptor}. GraphRAG extracts an
entity graph, detects communities, and pre-generates community summaries for global
queries~\citep{graphrag}. HippoRAG builds an open knowledge graph and runs
Personalized PageRank over it for single-step multi-hop retrieval, with HippoRAG~2
deepening passage integration toward non-parametric continual
learning~\citep{hipporag,hipporagtwo}. Zep maintains a temporal knowledge graph,
time-stamping and versioning facts so that updates and contradictions are reconciled
by recency rather than overwritten~\citep{zep}. Of the family it preserves
reversibility best: a superseded fact is retired, not deleted. The parametric
counterpart edits weights directly. ROME locates and rewrites the single mid-layer
MLP module that mediates a factual association via a rank-one update, and MEMIT
scales this to thousands of edits spread across a range of layers at
once~\citep{rome,memit}. Parametric editing eliminates retrieval latency, since
$\Uop$ becomes a forward pass, but it is maximally irreversible and hard to audit,
and at scale it induces the catastrophic forgetting that is the parametric analogue
of summary collapse.

EM-LLM collapses the distance between this layer and the KV layer of
Section~\ref{sec:kv}~\citep{emllm}. It segments the token stream into events by
surprise, stores each event's KV block as an episodic memory unit, and at decode time
retrieves relevant events by similarity and temporal contiguity, a retrieval over a
memory store whose items happen to be KV blocks rather than text notes. It is at once
a KV-cache method (the substrate is on-device key-value tensors) and an agent-memory
method (the operator is event segmentation plus salience retrieval), which is why it
tests the bridge thesis empirically: the same reversible, retrieval-backed design
that keeps an agent's note store from collapsing keeps an episodic KV memory from
collapsing, because both obey Eq.~\eqref{eq:bound} with the same $\Istar(Q)$.

Persistent, self-written, and trusted on later retrieval, $\Zrep$ is also an attack
surface. AgentPoison backdoors an agent by poisoning its long-term memory or RAG
store so that a trigger in a future query retrieves the malicious entry and steers
the agent's action~\citep{agentpoison}. MINJA injects a corrupted memory through
query-only interaction, with no privileged access, exploiting the very
write-on-experience loop that consolidation relies on~\citep{minja}. These are the
adversarial form of the self-reinforcing-error problem: a single planted memory, like
a single bad reflection, biases all subsequent retrieval, and reversibility offers
little defense once the poisoned item is the one retrieved.

\paragraph{Reversibility and the compounding of loss.} Across all four families the
distortion surrogate is LLM-judged salience, and (P-rev) is the axis that decides
their fate. Reversible schemes (archival paging with raw copies, temporal graphs,
multi-fidelity trees) can re-derive a dropped detail when a later query needs it, so
their error stays near the budget-limited floor of Eq.~\eqref{eq:bound}. Irreversible
schemes (lossy summarization, fold/prune, UPDATE/DELETE, weight edits) cannot, and
applied repeatedly, each compaction event composes its loss with the last. This is
prediction~(4): under repeated irreversible summarization, end-task error should grow
super-linearly in the number of compaction events, because errors both accumulate and
self-reinforce (a stored mistake biases the retrieval that feeds the next summary),
whereas a reversible, retrieval-backed memory stays flat. It is also the layer's
least-measured failure. Benchmarks such as LOCOMO and LongMemEval probe single-shot
recall, multi-session reasoning, and knowledge updates, but not the compounding curve
over many consolidation cycles~\citep{locomo,longmemeval}. That gap is why we test it
directly in Section~\ref{sec:exp}, and why recent surveys of agent memory and its
security flag lossy consolidation and memory integrity as the open
problems~\citep{surveyagentmem,surveyagentsec}.

\section{Trainable Sparse Attention}
\label{sec:sparse}

The KV-cache methods of Section~\ref{sec:kv} share a structural limitation that the
formalism names precisely. They are training-free heuristics that fix a compaction
operator $\Cop$ after the model is trained, then ask the frozen weights to tolerate
it. On the learnability axis (Section~\ref{sec:taxonomy}) they sit at the
``training-free'' end; on the query-adaptivity axis most sit at ``query-agnostic.''
That double placement is the exact corner the lower bound penalizes. An evictor that
decides what to keep before seeing the decode-time query pays the $H(Q)$ price of
consequence~(iii) in Section~\ref{sec:formalism}, and a heuristic surrogate
(accumulated attention mass, recency) is only a coarse estimator of the true
distortion. The methods here attack both deficits at once, making $\Cop$ itself
learned and end-to-end: the sparsity pattern is a parameter the model is trained
around, so the weights and the retain/discard rule co-adapt instead of fighting.
This is the learnability$\times$query-conditioning corner that no frozen-model evictor
can reach, and where the 2025 production frontier has settled, spanning both the KV
layer and the architecture itself.

\paragraph{Natively trainable sparse attention.} These methods train with sparse
attention from the start rather than retrofit it. Native Sparse Attention
(NSA)~\citep{nsa} composes three branches whose outputs are gated and summed:
coarse-grained compression of token blocks into block-level summaries, fine-grained
selection of the most relevant blocks for full attention, and a sliding window for
local context. The selection is differentiable and the kernels are hardware-aligned
to GPU tensor cores, so the scheme is both trainable end-to-end and fast on real
hardware, matching or beating full attention while reading a fraction of the cache.
In our terms NSA is a learned, query-conditioned $\Cop_\theta$ whose compression
branch supplies a low-rate global sketch and whose selection branch supplies a
high-fidelity tier, a multi-fidelity profile (P-fid) the trained weights learn to
exploit. MoBA~\citep{moba} reaches the same corner through MoE-style routing: a
learned gating function routes each query to a sparse subset of key--value blocks, so
a query attends only to the blocks a trained router deems relevant, and the model can
switch between sparse and full attention without retraining. The retain decision is
thus query-conditioned by construction, and the router is the query-aware selector
that Quest approximates heuristically. DeepSeek Sparse Attention (DSA)~\citep{dsa},
shipped in DeepSeek-V3.2, makes the frontier concrete: a lightweight ``lightning
indexer'' scores which past tokens each query should attend to, and fine-grained
selection over those tokens cuts long-context cost in a deployed model. DSA shows
that learned sparse attention has crossed from research into production
infrastructure.

A complementary line keeps the model weights fixed and instead learns, or calibrates
offline, the shape of the sparsity, exploiting the fact that attention maps are
structured. MInference~\citep{minference} finds that long-context prefill attention
falls into a few recurring patterns (A-shape, vertical-slash, and block-sparse),
assigns each head its pattern offline, then computes only the dynamically located
nonzero regions at inference, slashing the quadratic prefill cost with little quality
loss. This is a query-conditioned $\Cop$ acting at the prefill stage of the lifecycle
axis, the one stage the frozen decode-time evictors of Section~\ref{sec:kv} leave
untouched. Even without retraining, then, recognizing the learned structure of
attention recovers much of what a query-agnostic heuristic discards.

Attention heads are not interchangeable: a minority of ``retrieval heads'' carry the
long-range copying that needles depend on, while the rest are effectively local.
DuoAttention~\citep{duoattention} learns a binary label per head with a lightweight
optimization. Retrieval heads keep the full KV cache; streaming heads keep only
attention sinks plus a recent window. The expensive full-fidelity budget thus goes
only where the bound says it must. RazorAttention~\citep{razorattention} reaches a
similar split training-free, identifying retrieval heads by their attention structure
and compressing the rest while protecting dropped information with a compensation
token. Both instantiate prediction~(1): full fidelity is preserved on exactly the
heads whose evicted tokens would otherwise cause needle recall to collapse, and
budget is reclaimed from heads whose answer-relevant information content $\Istar$ is
low. They differ on the learnability axis, DuoAttention learning the label and
RazorAttention inferring it, but both allocate the same fidelity budget at head
granularity.

SeerAttention~\citep{seerattention} makes the selection rule an explicit learnable
component: a self-distilled gate learns each block's attention sparsity intrinsic to
the model, turning the implicit sparsity that evictors guess at into a trained,
queryable signal. Two other methods make the budget structure-aware.
SCOPE~\citep{scope} observes that prefill and decode have different compaction needs,
since decoding accumulates its own heavy hitters, and allocates separate KV budgets to
the two phases with a sliding decode strategy, dropping the assumption built into
decode-time evictors that one budget fits both stages of the lifecycle.
SepLLM~\citep{sepllm} exploits a structural regularity: separator tokens absorb
disproportionate attention mass, so a segment's information can be compressed into its
trailing separator, a structural token-level summarization that applies training-free
and can also be folded into training.

These methods answer the query-agnostic failure diagnosed in Section~\ref{sec:kv}
head-on. By folding $\Cop$ into the trained model, NSA, MoBA, and DSA make the
retain/discard decision query-conditioned and co-adapted with the weights,
sidestepping the $H(Q)$ penalty instead of paying it. MInference recovers the prefill
stage, DuoAttention and RazorAttention spend fidelity by head as the bound
prescribes, and SeerAttention learns the gate, while SCOPE makes the budget
phase-aware and SepLLM the summary structure-aware. The cost is the learnability itself: a frozen
evictor is plug-and-play, whereas natively sparse attention demands training or
distillation and so binds the operator to a model family. The direction of travel is
plain. The compaction operator is migrating out of the post-hoc heuristic layer and
into the model, which is why these methods, unlike those of Section~\ref{sec:kv}, are
the ones that reach production.

\section{Multimodal and Multi-Agent Compaction}
\label{sec:multi}

Two settings break the assumption that the unit of compaction is a text token in
a single model's context. A multimodal model's history $\Hist$ is dominated by
visual tokens; a multi-agent system's history is a collection of
inter-agent messages spread across collaborators. Both fight the same
quadratic-attention bottleneck, yet they sit at very different points of the
rate--distortion picture of Section~\ref{sec:formalism}, because their redundancy
structures differ. The formalism predicts where. At fixed quality the achievable
budget scales with the task-conditioned information content $\Istar(Q)$, and a
modality whose tokens are mutually predictable carries a small $\Istar(Q)$ per
token, so it survives much harder compression than text before the bound bites.

\paragraph{Compacting the visual stream.} A single image expands into
hundreds of patch tokens whose neighbors are spatially correlated, and video
multiplies this again across near-duplicate frames. The mutual information
$\I(\Zrep;\Hist)$ needed to preserve answer-relevant content is therefore a small
fraction of the raw token count, and in practice an $8$--$18\times$ reduction costs
almost nothing. The earliest responses were architectural. A Perceiver Resampler lets a small set of learnable latent queries cross-attend over
variable-length image or video features and distill them into a fixed budget of,
say, $64$ visual tokens~\citep{flamingo}. This is a learned $\Cop$ with a hard,
query-agnostic rate cap. It decouples LLM cost from input resolution, but because it
ignores the query it pays the $H(Q)$ penalty of Eq.~\eqref{eq:obj} and caps
fine-grained detail on dense or OCR tasks. Training-free methods soon exploited the
same redundancy with no retraining. Token Merging (ToMe) gradually merges the most
similar tokens in each block via bipartite soft-matching on key cosine
similarity~\citep{tome}; its merge operator is irreversible (P-rev) and its
distortion surrogate is pure embedding similarity, so it is task-agnostic and tends
to over-merge distinct foreground. FastV keeps full attention in the first two LLM
layers and then prunes low-attention visual tokens in deeper layers~\citep{fastv},
exploiting that visual tokens turn redundant after early layers. Its surrogate is
text-to-visual attention mass, which later work showed misaligns with true visual
importance, exactly as the bound's importance-heuristic analysis predicts.
VisionZip moves the compression ahead of the LLM, selecting dominant tokens by
encoder attention and merging the rest into a small contextual
set~\citep{visionzip}. LLaVA-PruMerge instead reads CLS-to-patch attention sparsity
through an IQR outlier criterion, prunes adaptively, and re-merges the pruned tokens
back into the survivors to reach $\sim18\times$ compression~\citep{prumerge}. ToMe, VisionZip, and LLaVA-PruMerge are query-agnostic (P-q absent) and irreversible
(P-rev absent); FastV is only weakly query-conditioned through its text-to-visual
attention surrogate but is likewise irreversible. The bound pins their shared
failure mode: evidence for a later, unanticipated grounding query is merged away
once and cannot be recovered, which is just the detail-sensitive regime where these
methods break down.

A second line carries the
KV-cache machinery of Section~\ref{sec:kv} over to the mixed-modality cache. Its
governing fact is that the two modalities have different $\Istar$ per token, which
makes a uniform budget the wrong default. LOOK-M evicts and merges KV entries under
a text-prioritized policy, protecting textual keys while merging the redundant
visual ones~\citep{lookm}; this encodes the observation that text tokens carry more
answer-relevant bits per token than image tokens do. VL-Cache turns the same
asymmetry into explicit budget allocation, distributing a global cache budget across
layers and modalities according to their measured sparsity~\citep{vlcache}. A
modality-aware $\budget$ split is the right move whenever one stream's tokens are
far more compressible than the other's.

A fixed cache is hopeless once the stream
runs to $10$K+ frames, so systems tier short- and long-term memory, an architectural
analogue of the agent-memory consolidation of Section~\ref{sec:agent}. MovieChat
fills a FIFO short-term buffer with dense frame tokens and, on overflow, merges
adjacent similar embeddings into a compact long-term memory~\citep{moviechat}. Its
Atkinson--Shiffrin-style two-tier $\Cop$ consolidates by similarity and ignores the
question, so query-specific evidence can vanish on overflow. MC-ViT takes a
different route, fine-tuning a pretrained video transformer to attend to a
non-parametric memory bank of redundancy-reduced past-segment
activations~\citep{mcvit}; this extends temporal context without architectural
change, but the memory still grows with video length. Both are multi-fidelity in
spirit, pairing a dense recent tier with a consolidated distant tier (P-fid), and
both stay irreversible. The bound warns that without query-conditioning the
consolidation step discards bits the future may need, and needle-in-a-long-video
recall is indeed where these methods are weakest.

When the history is spread across
collaborating agents, the unit becomes the inter-agent message, and the redundancy
is discrete-semantic, repeated reasoning and restated subgoals rather than
continuous spatial structure. Compaction turns abstractive here: it summarizes
transcripts into reusable insights, and it carries provenance and access control
that single-context compaction never has to. G-Memory builds a hierarchical
three-tier graph, an interaction-transcript graph, a query graph, and an insight
graph, that distills raw inter-agent dialogue into condensed trajectories and
abstract, reusable insights retrieved by bi-directional traversal~\citep{gmemory}.
In the rate--distortion view this is a lossy semantic $\Cop$ over messages whose
distortion surrogate is the LLM's own salience judgment, as in
Section~\ref{sec:agent}, with one extra constraint: an edge encodes which agent may
read which fragment, so the budget is shared across principals instead of owned by
one.

The two threads barely touch. On the visual
side the work compresses continuous, spatially redundant embeddings, almost always
training-free and similarity-based; on the multi-agent side it abstracts over
discrete messages under governance. Their intersection has no information-theoretic
account: compacting multimodal context shared across collaborating agents, a
blackboard holding compressed video memory that several agents read under different
access rights. The bound of Eq.~\eqref{eq:bound} is stated for a single $\Zrep$
answering a single $Q$, and it says nothing yet about how to size a shared $\budget$
when agents draw different $\Istar(Q)$ from one compressed multimodal store, or how
irreversible visual merging collides with the provenance a downstream agent may
later need. We still lack a theory that quantifies how much visual and temporal
redundancy is compressible without task loss, set against the discrete-semantic
redundancy of messages, and such a theory is what would ground the robust empirical
fact that visual tokens compress far harder than text.

\section{The Inference\,$\leftrightarrow$\,Agent-Memory Bridge}
\label{sec:bridge}

The four communities surveyed so far act on different substrates at different
timescales, yet they are working the same design problem. That problem shows up at
three tiers of a single memory hierarchy, and because it is one problem, the
mechanisms each community has refined carry across tiers rather than staying put.
This section makes the claim concrete: it aligns the tiers, traces the same five
design knobs through each of them, ports specific mechanisms across the bridge,
marks where the analogy breaks down, and ends with five compaction-aware design
principles the formalism implies.

\subsection{One hierarchy, three tiers}

A modern agent keeps memory at three scales that the literature usually studies
apart. The \textbf{KV cache} operates sub-second, on-GPU, and near-lossless at the
granularity of a token: it holds the attention state of the current forward pass,
and compaction here (Section~\ref{sec:kv}) decides which tokens' keys and values
survive within a single generation. The \textbf{working context} lives inside the
context window for the span of a task: it holds the agent's running trajectory,
and compaction here (Section~\ref{sec:agent}) decides which observations, tool
results, and reasoning steps stay visible as the task unfolds. The \textbf{long-term
store} is external and semantic and persists across sessions: it holds what the
agent carries from one task to the next, and compaction here decides which facts,
summaries, and episodes get written, kept, and retrieved. Their timescales span
many orders of magnitude, but all three answer the same Eq.~\eqref{eq:obj}: keep
the information a future query will need, under a budget, at minimal distortion.

\subsection{The same five knobs}

Since the three tiers share one objective, the same design knobs turn up at each,
and a knob one community has studied in depth is often blank in another
(Table~\ref{tab:bridge}). The importance signal is attention score in the KV
cache~\citep{h2o} and LLM-judged salience in the store~\citep{mem0}; both are
surrogates for the same distortion gradient (Section~\ref{sec:formalism}).
Forgetting takes the form of a sliding window or eviction schedule in the
cache~\citep{streamingllm} and an explicit decay curve in agent
memory~\citep{memorybank}. Query-conditioning divides methods that keep all state
and select per query (Quest;~\citealp{quest}) from those that commit in advance,
with retrieval at recall time versus eager summarization as the agent analogue.
Reversibility runs from keep-all paging to permanent discard at every tier. The
last knob, a calibrated stop rule for when compaction has gone too far, exists at
the KV layer as an output-error bound~\citep{adakv} and is all but absent for
agent summarization, a gap the bridge brings into view.

\begin{table}[t]
\centering
\small
\setlength{\tabcolsep}{3.0pt}
\renewcommand{\arraystretch}{1.25}
\begin{tabular}{@{}p{1.5cm} p{2.05cm} p{2.75cm}@{}}
\toprule
\textbf{Design knob} & \textbf{KV cache} & \textbf{Agent long-term memory} \\
\midrule
Importance signal & attention score \citep{h2o,snapkv} & LLM salience \citep{mem0,generativeagents} \\
Forgetting        & window / eviction \citep{streamingllm} & decay curve \citep{memorybank} \\
Query-cond.\ (P-q) & per-query pages \citep{quest} & recall-time retrieval \citep{raptor} \\
Reversibility (P-rev) & keep-all vs.\ evict \citep{quest,h2o} & archive vs.\ summarize \citep{memgpt} \\
Budget alloc.     & per-head/layer \citep{pyramidkv,adakv} & per-memory-type \citep{mirix} \\
Stop rule         & output-error bound \citep{adakv} & \emph{(open)} \\
\bottomrule
\end{tabular}
\caption{The bridge. The same design knobs recur across the memory hierarchy; a
mechanism refined at one tier is a candidate design for another, and an empty cell
(a principled stop rule for agent summarization) is a research opportunity the
alignment makes visible.}
\label{tab:bridge}
\end{table}

\subsection{Mechanisms that port across the bridge}

Once the knobs are read as shared, each column becomes a source of designs for the
others. Three ports show the payoff. MemoryBank's Ebbinghaus-style forgetting
curve~\citep{memorybank}, a recency- and reinforcement-weighted decay for
cross-session memory, is the content-aware prior KV eviction has been missing: H2O
and its successors evict by accumulated attention alone~\citep{h2o}, a signal
biased toward early tokens, whereas a decay prior calibrated to re-access
statistics drops straight in as a scoring alternative. Quest's discipline of never
evicting, and instead selecting query-relevant pages on demand~\citep{quest},
gives agent memory a template: an agent that retrieves from an archival store at
recall time (reversible and query-conditioned) dominates one that eagerly
summarizes, on exactly the queries the summary dropped. Ada-KV's output-error
bound~\citep{adakv}, which caps the deviation a given eviction induces in the
attention output, supplies the stop rule agent summarization lacks; instead of
compacting on a fixed schedule, an agent could compact until a bounded estimate of
induced output error is hit. EM-LLM~\citep{emllm}, which segments the KV cache into
episodic events and retrieves them like memories, already sits on the bridge and
is the cleanest empirical sign that the two tiers run on one logic.

\subsection{Where the analogy breaks}

The analogy has real limits, and they are worth stating. Recall guarantees differ
across the tiers: a KV cache that keeps a token's entries can attend to it exactly,
while a store that keeps a summary of an episode has already paid an irreversible
distortion. The meaning of ``query'' differs too. At the KV tier it is the
next-token computation, implicit in the current hidden state; at the agent tier it
is a future user request, genuinely unknown and possibly adversarial. Cost is the
third difference. KV compaction is a microsecond arithmetic decision, whereas agent
consolidation is itself an expensive, error-prone LLM call, so the agent tier pays
a budget the cache does not: the cost of deciding what to keep. None of this
refutes the bridge; it refines it, telling us which ports are clean (scoring
signals, reversibility disciplines) and which need adaptation (anything that
assumes free or exact recall).

\subsection{Five compaction-aware design principles}

Between them, the formalism and the bridge imply five principles for systems that
compact memory well. Each comes with supporting evidence and a failure the
formalism predicts when the principle is ignored.

\textbf{P1. Never discard irreversibly what you cannot re-derive cheaply.}
At equal budget a reversible operator (retrieval-backed eviction, archival memory)
weakly dominates an irreversible one (permanent eviction, lossy summarization): the
two can differ only on queries whose evidence was dropped, and there the reversible
one wins. Quest~\citep{quest} follows P1; H2O~\citep{h2o} and fixed-schedule
summarization violate it, with the predicted failures of recall collapse
(prediction~1) and error accumulation (prediction~4).

\textbf{P2. Condition on the query distribution, not just the current query.}
The bound charges a query-agnostic operator the entropy $H(Q)$
(Section~\ref{sec:formalism}). Trainable sparse attention
(Section~\ref{sec:sparse}) and query-aware compression~\citep{longllmlingua} pay
less of this penalty; offline gisting pays it in full.

\textbf{P3. Separate a cheap reversible episodic tier from a lossy semantic
tier, with explicit promotion and demotion.} Multi-fidelity memory (P-fid)
dominates a single lossy tier: keep raw episodes cheaply and recoverably, and
abstract only what proves reusable. MIRIX's typed modules~\citep{mirix} and the
Compressive Transformer's dual tier~\citep{compressivetransformer} are built this
way; a flat summarization buffer is not.

\textbf{P4. Make compaction asynchronous and learnable, not a blocking
heuristic.} When deciding what to keep is itself an LLM call on the critical path,
it erodes the savings it buys. RL-learned policies~\citep{contextfolding,memact}
and sleep-time consolidation~\citep{sleeptime} move the decision off the critical
path and learn it; fixed-threshold auto-compaction does neither.

\textbf{P5. Budget by marginal task-utility, not a uniform ratio.} The bound is
governed by $\Istar(Q)$, which spreads unevenly across heads, layers, and memory
items. Non-uniform allocation~\citep{pyramidkv,adakv} respects this; a global fixed
ratio ignores it and spends budget on low-utility state.

\begin{figure}[t]
\centering
\begin{tikzpicture}[>=Latex,
  tier/.style={draw,rounded corners,minimum width=6.7cm,minimum height=10mm,align=center,font=\small}]
\node[tier,fill=blue!10](kv){\textbf{KV cache}: sub-token, on-GPU\\[1pt]\scriptsize microseconds, near-lossless, one forward pass};
\node[tier,fill=teal!10,below=3mm of kv](wc){\textbf{Working context}: tokens, in-window\\[1pt]\scriptsize milliseconds to seconds, within a task};
\node[tier,fill=orange!12,below=3mm of wc](lt){\textbf{Long-term store}: semantic items, external\\[1pt]\scriptsize hours to days, across sessions};
\draw[<->,thick,gray] (kv) -- (wc);
\draw[<->,thick,gray] (wc) -- (lt);
\node[below=2.5mm of lt,text width=7cm,align=center,font=\scriptsize]{%
Shared design knobs (Table~\ref{tab:bridge}): importance scoring, budget
allocation, query-conditioning, forgetting, reversibility.};
\end{tikzpicture}
\caption{The memory hierarchy of Section~\ref{sec:bridge}. Three tiers separated by
many orders of magnitude in timescale share the same five design knobs, so a
mechanism refined at one tier becomes a candidate design for the others.}
\label{fig:hierarchy}
\end{figure}
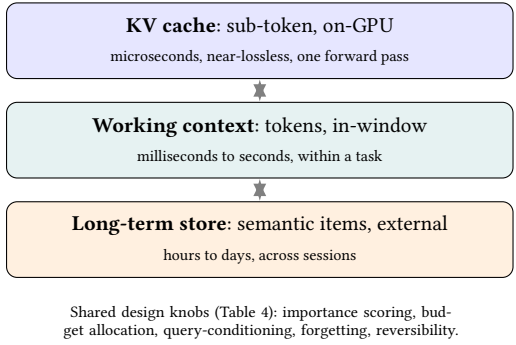
\section{Systems and Serving Substrate}
\label{sec:systems}

The preceding sections treated compaction as an algorithm. This one treats it as a
resource policy. At the serving runtime, the lifecycle stage in our
taxonomy (Section~\ref{sec:taxonomy}) where memory is managed across requests
rather than within one forward pass, the KV cache stops being a per-sequence
buffer and becomes a shared, multi-tier asset whose allocation, reuse, and
movement a scheduler decides. The same rate--distortion objective
(Eq.~\ref{eq:obj}) still governs, but the currency of $\rate(\Zrep)$ is now GPU
high-bandwidth memory weighed against bus bandwidth and latency SLOs, and $\Cop$
lives in a memory manager rather than an attention kernel. Read this way, serving
systems split into two families that the lens places at opposite ends of the
(P-rev) axis, and the interaction between them is the central tension of the stack.

The first family does not compress at all; it retains and shares exact KV.
PagedAttention~\citep{pagedattention} ports OS virtual-memory paging to the cache,
storing it in fixed-size non-contiguous blocks behind a per-sequence block table.
This eliminates fragmentation and, decisively for our argument, enables
copy-on-write sharing of identical prefixes across requests. RadixAttention in
SGLang~\citep{radixattention} generalizes the idea to a radix tree keyed by token
sequence, so an incoming request automatically reuses the longest cached prefix:
system prompts, few-shot exemplars, and branching reasoning trees are computed
once and read many times. CachedAttention~\citep{cachedattention} stretches the
horizon across conversational turns, spilling KV to DRAM and SSD tiers with
layer-wise prefetch to hide slow-tier latency. vAttention~\citep{vattention}
questions paging itself, using CUDA virtual-memory APIs to keep the cache
contiguous, so unmodified kernels run, while still allocating physical pages on
demand. In our formalism these are the limiting case of zero distortion: a cached
prefix is perfectly reversible memory, $\Cop$ approaching the identity, the
opposite extreme from eviction. Their leverage comes not from spending fewer bits
but from amortizing bits already spent, which is why the gains are entirely
workload-dependent and collapse to nothing without prefix overlap. Reuse, unlike
compression, buys nothing when there is nothing to reuse.

A second runtime concern is not how large the cache is but where it lives. Prefill
is compute-bound and decode is memory-bound, so co-locating them lets each
interfere with the other's SLO. Splitwise~\citep{splitwise},
DistServe~\citep{distserve}, and Mooncake~\citep{mooncake} disaggregate the two
phases onto separate machines, which turns the KV cache into a tensor that must be
transferred between prefill and decode pools over the interconnect.
DistServe places the phases by cluster bandwidth to bound transfer cost while
optimizing TTFT and TPOT independently for goodput. Mooncake goes further and pools
idle CPU, DRAM, and SSD across the cluster into a disaggregated KVCache store driven
by a cache-centric scheduler. Here $\rate(\Zrep)$ returns not as an occupancy budget
but as a bandwidth budget: the cost of an exact, reversible representation is the
time to ship it, which is what makes the lossy and the transmissible families below
attractive even when memory is not scarce.

The third family reintroduces distortion to fit large models and long contexts into
a fixed hierarchy. FlexGen~\citep{flexgen} aggregates GPU, CPU, and disk, solves a
linear program for tensor placement, and quantizes weights and the cache to four
bits, trading precision and interactivity for capacity in offline batch settings.
InfiniGen~\citep{infinigen} keeps only a working set on-GPU, rehearsing the next
layer with partial weights to guess which few tokens will matter and prefetching
only their KV from host memory. ShadowKV~\citep{shadowkv} offloads the value cache,
retains a low-rank key cache plus landmarks and outliers, and reconstructs keys to
fetch only the values it needs. These are the query-conditioned, approximate
operators of Section~\ref{sec:kv}, now embedded in the memory manager: their
selection is lossy (P-fid) and their dropped entries are not re-derivable within the
request (irreversible, $\neg$P-rev).

A bridging family makes exact KV cheap to move rather than cheap to
store. CacheGen~\citep{cachegen} encodes the cache into a compact bitstream that
exploits its layer- and token-locality and streams it, trading a controlled,
near-lossless quantization error for far lower transfer time, which puts compaction
in the service of disaggregation. CacheBlend~\citep{cacheblend} attacks a limitation
all of the above share, that they reuse only prefixes: in RAG the relevant chunks
arrive concatenated in query-dependent order, so prefix matching fails. CacheBlend
reuses each chunk's independently precomputed KV and recomputes only the small
fraction of cross-attention entries that the chunks' new neighbors change,
recovering most of the quality of full attention at a fraction of the prefill cost.
This is a multi-fidelity operator (P-fid) in the wild: a large reused tier plus a
tiny recomputed tier sized by an error budget.

The lens exposes a tension these systems must resolve jointly. Lossless reuse
(PagedAttention, RadixAttention, prefix caches) assumes the stored KV is exactly
what attention would have produced; lossy budgeting (InfiniGen, ShadowKV,
quantized offload) assumes it may be approximated. Composing the two operators
costs something: reusing a prefix that was itself evicted or quantized propagates
the earlier distortion into every request that shares it, and the errors compound
across the reuse tree instead of staying local to one sequence. With the two
families at opposite ends of (P-rev), the open systems problem is a scheduling
question over one budget. When should KV stay reversible and exact, paying storage
and bandwidth, and when should it go lossy and cheap, paying distortion? And how
does one stop a lossy decision made for a single request from silently degrading
every descendant that later reuses it?

\section{Theory of Compaction}
\label{sec:theory}

The formalism of Section~\ref{sec:formalism} reduces compaction to a single
rate--distortion program~\eqref{eq:obj} with a single lower bound~\eqref{eq:bound}.
This section collects the rigorous results that either instantiate that bound or
sharpen the constant in front of it, and closes on the one quantity the theory
still cannot predict: $\Istar(Q)$ itself.

The cleanest instance of the bound is a space lower bound on lossless KV caching.
\citet{compressionbarriers} prove that any algorithm performing exact
attention-based generation over a length-$n$, dimension-$d$ context must use
$\Theta(nd)$ space, with a matching $\Omega(d\cdot e^d)$ barrier in the
low-dimensional regime and no non-adaptive sublinear-time online attention. Read
through~\eqref{eq:bound}, this says exact attention demands $\budget \ge \Istar(Q)$
for the worst-case query: when the task is verbatim recall over an incompressible
context, $\Istar(Q)$ is itself $\Theta(nd)$, so no $\Zrep$ smaller than the cache
can drive $P_e$ to zero. Sub-linear caches are therefore impossible without
structural assumptions on $\Hist$, the very assumptions our $1/\Istar(Q)$
compression ratio exploits whenever real data is redundant. The same shape recurs
for fixed-state recurrence. \citet{rnnsnottransformers} show that fixed-size RNN/SSM
states provably cannot solve the in-context associative recall or tree recognition
that attention solves, that chain-of-thought alone does not close the gap, and that
a single attention layer (or RAG) restores it. This is consequence~(iii) of our
bound made architectural: a state of $\budget$ bits cannot answer a query needing
more than $\budget$ bits, so the separation turns on exact in-context retrieval, the
regime where $\Istar(Q)$ is large and query-agnostic ($H(Q)$-blind) compaction pays
most. Parametric memory completes the picture. \citet{factualrecall} characterize it
directly: transformers store facts as associative memories whose capacity scales
linearly in parameter count, with a single attention-plus-MLP block near-optimal.
This puts a concrete $\Istar$-budget on the weights rather than the cache, and it
explains why some history need never be retained as $\Zrep$ at all, since facts
already amortized into parameters contribute zero to $\I(Y;\Hist\mid Q)$.

The bound says how many bits must survive; attention sinks say which tokens hold the
survival machinery. \citet{streamingllm} observed that softmax forces large
attention mass onto the first few tokens irrespective of content, and that retaining
$\sim$4 sink tokens plus a sliding window streams to millions of tokens.
\citet{attnsinkemerges} show empirically that sinks are universal, emerge only after
enough optimization, act as ``key biases'' that absorb surplus attention without
contributing to value computation, and vanish under un-normalized (sigmoid)
attention. A sink is therefore an artifact of the softmax normalizer, not of
$\Istar(Q)$. The decisive result for compaction is~\citet{sinknecessity}: for a
natural family of trigger-conditional tasks, any softmax model that solves them must
develop a sink, whatever its training. Sinks then carry irreducible $\Istar(Q)$ for
those tasks, so evicting them is not lossy compression but a violation of budget
feasibility in~\eqref{eq:obj}, and the failure is structural rather than graceful.
\citet{sinkvalleys} unify sinks with depth-localized ``compression valleys'' (sharp
drops in residual-stream rank) as two faces of the same massive-activation mechanism
concentrated in middle layers. On the representation side this predicts where
$\Hist$ is most compressible across depth, and it warns that the high-fidelity tier
of any (P-fid) profile must go to the anchor tokens the valleys form around.

The bound is also location-dependent.
\citet{lostinmiddle} document a U-shaped recall curve: accuracy is highest when the
evidence sits at the context's start or end and collapses in the middle. In our
terms this is a distortion $\dist(\theta)$ that depends on where the answer-bits sit
within $\Hist$, even at fixed $\budget$ and fixed $\Istar(Q)$. It sharpens
prediction~(1) of the formalism, since the eviction collapse point tracks the
entropy of the answer's location, so a compaction scheme uniform in position
silently raises $P_e$ for mid-context evidence. Placement, not retention alone, is a
free variable in the rate--distortion program.

The objective~\eqref{eq:obj} rests on the rate--distortion/IB problem of
\citet{tishbyib}: trade $\I(\Zrep;\Hist)$ against $\I(\Zrep;Y\mid Q)$, keeping the
answer-predictive bits and spending nothing on the rest. \citet{quitox} make this
operational for context compression, recasting token selection as maximizing mutual
information with the query/answer under a token budget and using cross-attention as
the relevance estimator. This is the query-conditioned (P-q) reading of our
surrogate-distortion view: it estimates $\I(\Zrep;Y\mid Q)$ directly rather than
through the perplexity proxy, which is why an IB objective beats self-information
pruning by roughly the mutual-information gap of prediction~(2).

Every result above is worst-case or mechanistic, and none predicts the 80--93\% KV
reductions and $\sim$20$\times$ prompt compression that practice routinely achieves,
with quantization reportedly within $\sim$2.7$\times$ of the information-theoretic
limit. The reconciliation is consequence~(ii). Worst-case $\Istar(Q)$ is
$\Theta(nd)$, but on realistic data $\Istar(Q)$ is far smaller, and the achievable
ratio is $\propto 1/\Istar(Q)$. The gap between the lower bounds and observed
compressibility is thus the gap between worst-case and data-conditional $\Istar(Q)$,
and that gap is not yet characterized. What the theory still owes us is a predictive
compression scaling law: a closed-form estimate of $\Istar(Q)$, and hence of the
achievable ratio at fixed quality, as a function of model size, context length,
head/layer structure, and task type. The results we have, linear parametric
capacity, near-limit quantization, sink necessity, U-shaped recall, are scattered
estimators of pieces of $\Istar(Q)$. Unifying them into one law that predicts,
before compaction runs, how much memory a given model on a given task can shed is
the central unsolved problem this survey leaves open.

\section{Evaluation and the COMPACT-Bench Proposal}
\label{sec:eval}

Whatever a compaction method claims, the claim only holds up to the evaluation
that backs it. The formalism of Section~\ref{sec:formalism} pins down three
things such an evaluation has to measure: the accuracy reachable at a fixed
memory rate $\budget$; whether a gap between two schemes at equal $\budget$
traces back to reversibility (P-rev) and query-conditioning (P-q); and whether
the four predictions hold layer by layer. No current benchmark measures these
directly. Below we review the suites that exist, pin down what they miss, and
sketch a benchmark built to fill it.

\subsection{What exists}
The field started from one synthetic probe and grew along three axes. Synthetic
stress tests buy controllable length and difficulty. Needle-in-a-Haystack drops
a single fact at varying depth and plots a depth$\times$length
heatmap~\citep{niah}. RULER widens this to thirteen multi-key, multi-hop, and
aggregation tasks, finding that barely half of the models advertising $\geq$32K
actually survive at 32K~\citep{ruler}. BABILong buries bAbI reasoning facts in
book-length filler and reports that models exploit only $10$--$20\%$ of their
stated window~\citep{babilong}. Each of these has a natural reading in terms of
$\Istar(Q)$: single-fact NIAH carries little information and saturates, whereas
the multi-hop demands of RULER and BABILong push $\Istar(Q)$ up and surface the
Fano floor of Eq.~\eqref{eq:bound}. A second family of suites trades control for
ecological validity. LongBench covers 21 multitask datasets~\citep{longbench},
InfiniteBench drives average length past 100K with dependency-heavy
tasks~\citep{infinitebench}, and HELMET layers in model-based judging and
few-shot prompting to repair the noisy $n$-gram signals and base-model
incompatibility that plagued the earlier suites~\citep{helmet}. The third family,
interactive memory benchmarks, aims at the agent layer. LOCOMO scores QA,
summarization, and temporal reasoning over multi-session dialogues~\citep{locomo}.
LongMemEval separates five memory abilities (extraction, cross-session and
temporal reasoning, knowledge updates, and abstention) over $\sim$115K-token
histories, where it sees $\sim$30\% drops on sustained
interaction~\citep{longmemeval}.

Two efforts come closest to compaction itself. SCBench studies long-context
methods through a KV-cache-centric lens, spanning generation, compression,
retrieval, and loading in a shared-context, multi-request
setting~\citep{scbench}; KVPress~\citep{kvpress} supplies a harness that standardizes
KV compression so eviction and quantization methods all run behind one interface.
We treat both as direct precursors and build on them rather than around them.
NoLiMa adds a cautionary result: literal-match probes like NIAH systematically
flatter the model. Strip the lexical overlap between needle and query so that
recall has to pass through latent association, and accuracy falls far below the
NIAH heatmap~\citep{nolima}. Read through Section~\ref{sec:formalism}, NIAH
measures $\Istar(Q)$ only in the easy regime where the answer-bearing bits are
lexically marked, and NoLiMa shows that the bits compaction actually discards are
the unmarked ones.

\subsection{The missing benchmark}
For all this coverage, no suite measures compaction the way the formalism
requires, and it falls short in two related ways. The first is the budget axis.
KV work reports tokens or GPU bytes, prompt work reports a compression ratio,
recurrent work reports state dimensions, and agent work reports store size. The
units never line up, so the frontiers are incommensurable and each method's
budget is set by whoever wrote the paper. The second gap is more fundamental:
none of these suites measures the capabilities that, by the formalism, tell good
compaction from bad at equal $\budget$, namely whether a scheme knows what it
dropped, whether it can get it back, and whether its confidence reflects what it
kept. Post-compaction end-task accuracy is an integral over $\dist(\theta)$. It
averages away the lost bits instead of locating them, so it cannot say which
unseen query will fail, and P-rev and P-q, the properties prediction~(4) ties to
repeated compaction, never get scored at all.

Our response is \textbf{COMPACT-Bench}, an extension of SCBench~\citep{scbench}
and KVPress rather than a replacement: it keeps their task corpora and KV harness
and supplies the axes and task families they leave out. Three pieces make it up.

The first is a shared budget axis. Every method lands on one
bytes-per-token-of-history (BPT) scale, total retained memory in bytes over raw
history tokens, the layer-agnostic form of $\rate(\Zrep)$, and is reported as a
full accuracy-versus-budget frontier instead of one operating point. This makes
the Eq.~\eqref{eq:bound} floor visible on the page and the $1/\Istar(Q)$
task-dependence of Section~\ref{sec:formalism} measurable across layers.

The second piece is three task families that probe the capabilities end-task
accuracy hides. \textbf{Loss attribution} checks whether a method knows what it
dropped: after compaction, probe recall on a held-out set of history facts and
compare the method's own predicted retention against the truth, a direct test of
its distortion surrogate. \textbf{Reversibility} checks whether dropped content
can be re-derived on demand: issue a late query whose evidence was evicted and
measure recovery, which separates retrieval-backed and archival schemes (P-rev
holds) from eviction and summarization (where it does not). \textbf{Calibrated
compaction confidence} checks whether a method's stated confidence tracks its
real post-compaction accuracy, so that an agent can abstain or re-expand instead
of answering from a lossy state.

The third piece reports accuracy, latency, memory, and dollar cost together on
one sheet, because a frontier that ignores decode latency or store cost conceals
the very trade a deployment has to make. Since NoLiMa shows literal-match probes
inflate quality~\citep{nolima}, all retrieval-style items use latent-association
needles rather than lexical ones, so the frontier reflects the unmarked bits that
compaction destroys. The protocol's sharpest test is prediction~(4): plot
end-task error against the number of compaction events, and it should climb
super-linearly for irreversible summarization while staying flat for reversible
memory, a contrast no existing suite can draw. Section~\ref{sec:exp} runs a first
version of this protocol on the KV and agent layers.

\begin{table}[t]
\centering
\small
\setlength{\tabcolsep}{4pt}
\renewcommand{\arraystretch}{1.18}
\begin{tabular}{l l ccc}
\toprule
\textbf{Benchmark} & \textbf{Type} & \makecell{\textbf{Multi-}\\\textbf{budget?}} & \makecell{\textbf{Cross-}\\\textbf{layer?}} & \makecell{\textbf{Repeated}\\\textbf{compact.?}} \\
\midrule
NIAH~\cite{niah}            & synthetic   & \none & \none & \none \\
RULER~\cite{ruler}          & synthetic   & \half & \none & \none \\
BABILong~\cite{babilong}    & synthetic   & \half & \none & \none \\
LongBench~\cite{longbench}  & realistic   & \none & \none & \none \\
InfiniteBench~\cite{infinitebench} & realistic & \none & \none & \none \\
HELMET~\cite{helmet}        & realistic   & \half & \none & \none \\
LOCOMO~\cite{locomo}        & interactive & \none & \none & \half \\
LongMemEval~\cite{longmemeval} & interactive & \none & \none & \half \\
SCBench~\cite{scbench}      & KV          & \half & \half & \half \\
KVPress~\cite{kvpress}      & KV          & \full & \none & \none \\
\midrule
\textbf{COMPACT-Bench}      & all four    & \full & \full & \full \\
\bottomrule
\end{tabular}
\caption{Benchmarks for long-context and compaction evaluation against the
properties the lens of Section~\ref{sec:formalism} requires. \emph{Multi-budget?}
sweeps a memory-rate axis; \emph{Cross-layer?} compares KV, prompt, architectural,
and agent methods on one axis; \emph{Repeated-compaction?} measures degradation
under repeated compaction events. \full~holds; \half~partial; \none~absent.}
\label{tab:bench}
\end{table}

\section{A Reference Experiment}
\label{sec:exp}

It is one thing to argue that the layers share a budget axis and a failure mode,
another to watch it happen. So we ran two small studies on a single commodity
GPU, an NVIDIA RTX~4060 with 8\,GB, using \texttt{Qwen2.5-1.5B-Instruct}. We are
not chasing state of the art here; we want to demonstrate the methodology the
survey advocates, namely a common budget axis that makes heterogeneous methods
comparable (Section~\ref{sec:eval}) and a measurement of the repeated-compaction
regime the field leaves untested. The absolute numbers are tied to this small
model, but the result is the shape of the two curves, and both shapes carry over
to larger scale. Code and configurations accompany the survey.

Every method sits on bytes-per-token-of-history (BPT), the bytes of retained
memory state over the tokens of original context. The retained state is a KV
cache in the end, so all four mechanisms collapse onto one currency: token
eviction that keeps a fraction $f$ costs $f$ of the full per-token KV bytes,
$b$-bit quantization costs $b/16$ of it, prompt or summary compression to $m$ of
$n$ tokens costs $m/n$, and the full cache anchors the top. This is the axis
COMPACT-Bench standardizes, and it is what puts a KV evictor and a quantizer on
the same chart.

\subsection{Experiment 1: the unified accuracy--budget frontier}

The task is needle-in-a-haystack retrieval with natural filler text drawn from
Wikitext: we plant a unique key--value needle at controlled depths (10\% to 90\%)
in contexts of 2k to 8k tokens and ask for the value. We sweep six KV-compaction
methods from Section~\ref{sec:kv}, namely SnapKV~\citep{snapkv},
StreamingLLM~\citep{streamingllm}, TOVA~\citep{tova}, the Knorm and
expected-attention scorers, and a random-eviction control, each at five budgets.
Accuracy is averaged over depths, lengths, and three trials, with the full cache
anchoring the top at 100\% budget.

\begin{figure}[t]
\centering
\includegraphics[width=\columnwidth]{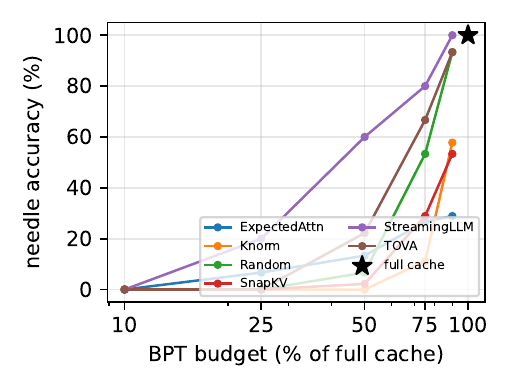}
\caption{The accuracy--budget frontier on natural-filler needle retrieval with
\texttt{Qwen2.5-1.5B}, six KV-compaction methods on one bytes-per-token-of-history
axis. Accuracy collapses once the budget falls below the task's information
content, and the random-eviction control isolates the value each scorer adds.}
\label{fig:frontier}
\end{figure}

Figure~\ref{fig:frontier} reports the result over 1{,}395 generations. The full
cache answers perfectly (accuracy $1.00$). As the budget falls, the methods trace
a frontier that slides toward zero, and below roughly a quarter of the full
budget every method sits at or near zero accuracy. Eq.~\eqref{eq:bound} predicts
exactly this: the needle task concentrates its task-conditioned information
content $\Istar(Q)$ in a few tokens, so once the budget drops under that content
no scorer can recover the answer, an instance of prediction~(1). The
random-eviction control does the diagnostic work. At high budget it stays
competitive, because keeping most tokens keeps the needle by luck, but it
collapses fastest as the budget tightens, so the gap between a real method and the
random line measures how much its scoring buys at a given budget. The gaps are
modest at this scale, and the ordering of methods is beside the point; what
matters is the comparability the BPT axis delivers.

\subsection{Experiment 2: error accumulation under repeated compaction}

No single-turn benchmark can run this test, which is precisely why it matters. An
agent reads a long document in chunks, accumulating twelve key--value facts spread
throughout, and periodically compacts its working memory. We compare two operators
as the number of compaction events grows. The irreversible operator overwrites
working memory with an LLM-generated summary. The reversible operator keeps all
chunks in an archive and retrieves the relevant chunk at query time, the Quest and
MemGPT-archival discipline of P1. At the end we ask for all twelve facts and
report recall.

\begin{figure}[t]
\centering
\includegraphics[width=\columnwidth]{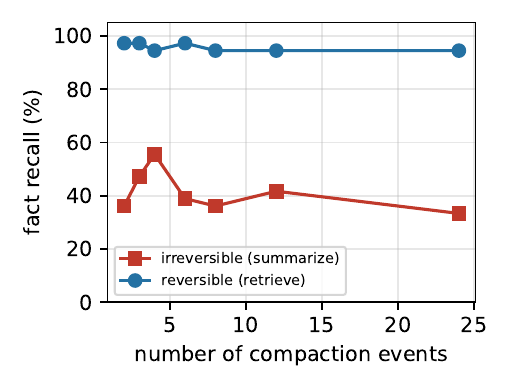}
\caption{Fact recall against the number of compaction events for an irreversible
(summarize) and a reversible (retrieve) operator. Reversible memory stays near
$0.95$; irreversible memory loses roughly half its facts at every frequency. The
two look identical under single-turn evaluation and diverge only when compaction
repeats.}
\label{fig:accum}
\end{figure}

Figure~\ref{fig:accum} shows the separation. The reversible operator holds recall
near $0.95$ at every compaction frequency, since retrieval can re-derive any
dropped fact. The irreversible operator runs far below it, between $0.33$ and
$0.56$, and is weakest at the highest compaction frequency, because each summary
throws away facts the next summary can no longer see and the loss compounds. Two
operators at the same average budget therefore part by roughly $0.5$ in recall
once memory is reused over a horizon, even though a single-turn needle test cannot
tell them apart. This is what it means in practice to say that at equal budget,
reversible compaction dominates.

Scale is not what makes either result land. The first shows that a single budget
axis renders heterogeneous compaction methods comparable, and that accuracy
collapses once the budget drops below the information content the task requires.
The second shows that the regime agents actually live in, repeated compaction over
a long horizon, splits reversible from irreversible operators that look identical
under single-turn evaluation. Each is a falsifiable instance of the formalism, and
each invites replication at larger scale.

\section{Open Problems and Research Agenda}
\label{sec:open}

The formalism of Section~\ref{sec:formalism} does more than unify the
literature: it tells us which problems are worth solving and why. We close by
setting out a prioritized agenda, ordered by impact times tractability. Each
item is stated against the bound~\eqref{eq:bound}, the three properties
(P-rev),(P-q),(P-fid), or the four predictions, so that ``progress'' has a
precise meaning rather than another point on a borrowed long-context curve.

First, query-conditioned, reversible, multi-fidelity compaction at every
layer. The single highest-leverage target follows directly from the
bound: a query-agnostic operator pays a price of $H(Q)$ bits it cannot recover
(consequence (iii)), and an irreversible one cannot re-spend its budget once the
query arrives. The agenda is to build operators that satisfy (P-q) and (P-rev)
simultaneously, with a (P-fid) profile that keeps a cheap lossy tier over
a small recoverable high-fidelity tier, and to do so at each tier of the
hierarchy, porting the query-conditioned, no-evict design of Quest~\citep{quest}
and the retrieval-backing of InfiniGen~\citep{infinigen} up into prompt and
agent memory. This is tractable because the components exist in isolation; the
work is composing them and showing the predicted right-shift (Prediction~2) and
flat error curve (Prediction~4).

Second, information-loss attribution and calibrated compaction confidence.
Every layer today measures only end-task accuracy after the fact and cannot say
what was dropped or predict failure on an unseen query. Because the
surrogate scorers of Section~\ref{sec:formalism} are all estimators of one
distortion, they can in principle emit a calibrated estimate of residual
$\Istar(Q)$, a per-decision ``compaction confidence'' with provenance of loss.
A method that knows it has fallen below $\budget < \Istar(Q)$ can defer, fetch,
or escalate fidelity; this attribution is the prerequisite for everything
adaptive below.

Third, a composition map for stacking operators. Quantization, low-rank
projection, eviction, and merging are treated as independent recipes, yet
stacking them compounds or cancels distortion, and serving-level lossless reuse
(RadixAttention;~\citealp{radixattention}) directly conflicts with lossy
budgeting (InfiniGen;~\citealp{infinigen}). We need an honest empirical
``operator algebra'', which pairs are synergistic, redundant, or destructive,
within a layer (quant $\times$ low-rank $\times$ evict) and across layers
(KV $\times$ prompt $\times$ agent), together with a treatment of the lossless-reuse
versus lossy-budget tension as a first-class trade. We stress that this is a
measured map, not a set of proven laws: distortion surrogates are not additive
and composition order matters.

Fourth, a predictive compression scaling law. The bound makes the
compression ratio at fixed quality scale as $1/\Istar(Q)$, but $\Istar(Q)$ is
never measured directly. The open problem is a predictive law that estimates
$\Istar(Q)$, and hence the achievable ratio, as a function of model size,
context length, head and layer structure, and task type, closing the gap between
worst-case lower bounds ($\Theta(nd)$, sink necessity) and the reported
80--93\% KV reductions. Such a law would turn budget selection from tuning into
calculation.

Fifth, an episodic-to-semantic consolidation bridge and a stop rule.
Within-task curation (ACON;~\citealp{acon}, Context-Folding;~\citealp{contextfolding})
and cross-session memory (Mem0;~\citealp{mem0}, MemGPT;~\citealp{memgpt}) use
disjoint mechanisms with no principled promotion from a lossless, recoverable
episodic tier to a lossy, abstracted semantic tier. The formalism supplies the
missing pieces: promote a span only when its marginal $\Istar$ over the query
distribution justifies abstracting it, and, because repeated irreversible
summarization grows error super-linearly (Prediction~4), adopt an output-error
stop rule for agent summarization in the spirit of Ada-KV's bound.

Sixth, cheap, asynchronous, RL-stable agent self-curation.
Reflection-, note-, and update-based memory is LLM-call-heavy and
self-reinforcing: it spends calls that erode the savings and stores errors that
bias all future retrieval. The target is curation that is asynchronous (off the
critical path, as sleep-time compute~\citep{sleeptime} begins to be), learned
rather than hand-triggered (MemAct;~\citealp{memact}, MEM1;~\citealp{memone}),
and stable under RL so the reward does not collapse into degenerate forgetting.

Seventh, multimodal $\times$ multi-agent shared-memory compaction. Visual
tokens are highly redundant and compress well (VisionZip;~\citealp{visionzip},
PruMerge;~\citealp{prumerge}); multi-agent memory carries provenance and
access-control constraints (G-Memory;~\citealp{gmemory},
MIRIX;~\citealp{mirix}). Nobody has compacted multimodal context shared
across collaborating agents, where the governing quantity is cross-modal versus
inter-message redundancy under per-agent query distributions, a clean
multi-source instance of objective~\eqref{eq:obj}.

Eighth, adaptivity with guarantees. Budget, bit-width, rank, and ratio
are allocated per layer, head, position, and turn by hand-tuned heuristics
(PyramidKV;~\citealp{pyramidkv}). The bound says the right allocation equalizes
the residual $\Istar(Q)$ across units, budgeting by marginal task-utility rather than
uniform ratio, so the open problem is content- and query-adaptive allocation
that comes with a guarantee on the resulting distortion rather than an empirical
sweep.

Ninth, safety, auditability, and rollback of evolving memory.
Self-editing memory that satisfies (P-rev) for utility also creates an attack
surface: poisoning, leakage, and irreversible drift (AgentPoison;~\citealp{agentpoison},
MINJA;~\citealp{minja}). Reversibility cuts both ways, enabling rollback and
audit but also persistent injection, so the agenda is auditable, version-able
compacted memory with a forgetting and revocation primitive, an area only
beginning to be governed (surveyed in~\citealp{surveyagentsec}).

Two threads run through this list. At equal budget, an operator that can re-derive
what it dropped (P-rev) and conditions on the query (P-q) beats one that discards
for good, and the bound says why: only the reversible, conditioned operator escapes
the $H(Q)$ penalty and the compounding error of repeated lossy summarization. And
the field still has no cross-layer benchmark; methods report on budgets they choose
for themselves over borrowed long-context suites, so their accuracy-versus-budget
curves cannot be compared across granularities. Until a common budget axis exists,
the agenda above will be measured in mismatched units, and the unification this
survey proposes stays a framing rather than a tested science.

\section{Conclusion}
\label{sec:conclusion}

We have argued that memory compaction in language models and their agents is one
problem wearing four disguises. Evicting a KV token, quantizing it, pruning a
prompt span, bounding a recurrent state, and summarizing an agent's trajectory are
all instances of a single rate--distortion decision: what context-derived
information to retain versus discard, at what fidelity, under a budget, so as to
preserve task utility (Section~\ref{sec:formalism}). Reading the whole stack through
this lens pays off three ways. It supplies a common currency: a budget axis on
which a KV evictor and an agent summarizer can finally be compared. It exposes a
single failure mode, the query-agnostic and irreversible loss of later-needed
information, recurring at every layer and predicted by the same bound. And it
licenses transfer: a forgetting curve becomes a KV prior, a no-eviction retrieval
discipline becomes an agent-memory design, and an output-error bound becomes a stop
rule for summarization (Section~\ref{sec:bridge}).

Two results stand out. At equal budget, a reversible operator beats an irreversible
one: of two methods that keep the same number of bytes, the one that can recover
what it dropped wins on the queries that depend on it, and the reference experiment
shows that gap widening as the memory is reused (Section~\ref{sec:exp}). And the
field has no benchmark that holds one budget axis across all four layers while
measuring repeated compaction; today's suites report ratios at fixed quality on
single-turn long context and leave the agentic regime, where compaction compounds,
untested. We give such a benchmark in Section~\ref{sec:eval} and run a first
version of it.

The agenda that follows (Section~\ref{sec:open}) is, in one sentence, to make
compaction query-conditioned, reversible, and attributable: to keep what a
future query will need, to recover what was dropped when it is needed
anyway, and to know what was lost. Progress will come not from a better heuristic
at any one layer but from recognizing that the layers share an objective and
measuring them the same way.

\section*{Limitations}

This is a survey: it proposes a formalism and a benchmark protocol but does not
provide large-scale empirical validation of either. The lower bound
(Eq.~\ref{eq:bound}) assumes a query-agnostic operator for its cleanest form and
treats the task-conditioned information content $\Istar(Q)$ as given; estimating
$\Istar(Q)$ for real tasks is itself open (Section~\ref{sec:theory}). The
cross-layer ``operator algebra'' we call for is, at present, an empirical
composition map rather than a set of proven composition laws. Our reference
experiment (Section~\ref{sec:exp}) runs at reference scale (a small open model on a
single commodity GPU), so its absolute numbers should be read as illustrations of
the methodology, not as a leaderboard; the trends it isolates (the accuracy--budget
frontier and error accumulation under repeated compaction) are the contribution,
and they invite replication at scale. Finally, the field moves quickly and many
cited works are recent preprints whose results await independent reproduction; we
have flagged these where they bear on our claims, and reported compression ratios
are drawn from heterogeneous setups that the normalization protocol of
Section~\ref{sec:eval} is meant to reconcile rather than take at face value.

\bibliographystyle{ACM-Reference-Format}
\bibliography{refs}

\appendix
\section{Master Method-by-Axis Table}
\label{app:master}

Table~\ref{tab:master} classifies representative methods from across the survey
along the seven axes of Section~\ref{sec:taxonomy}. Abbreviations.
Granularity: tok (token/KV entry), pg (page/block), bit (bit-width),
rk (rank), hd (head/layer), span (NL span), gist (soft vector), st (recurrent
state), fact (semantic item), vtok (visual token).
Lifecycle: pre (pretraining), pf (prefill), dec (decode), srv (serving),
task (within-task), x-task (cross-task).
Rev.: \full\ reversible, \half\ partial, \none\ irreversible.
Adapt.: agn (query-agnostic), qry (query-conditioned), lrn (learned reward).
Learn.: free (training-free), adpt (adapter/post-train), scr
(from scratch), RL, ctrl (LLM controller).

\begin{table*}[t]
\centering
\scriptsize
\setlength{\tabcolsep}{4pt}
\renewcommand{\arraystretch}{1.15}
\begin{tabular}{l l l l c l l l}
\toprule
\textbf{Method} & \textbf{Layer} & \textbf{Granularity} & \textbf{Lifecycle} & \textbf{Rev.} & \textbf{Adapt.} & \textbf{Learn.} & \textbf{Mechanism} \\
\midrule
StreamingLLM \citep{streamingllm} & KV & tok & dec & \none & agn & free & evict (sink+window) \\
H2O \citep{h2o}              & KV & tok       & dec   & \none & agn & free & evict (attn score) \\
SnapKV \citep{snapkv}        & KV & tok       & pf    & \none & qry & free & evict (obs.\ window) \\
PyramidKV \citep{pyramidkv}  & KV & tok/hd    & pf    & \none & agn & free & evict (layer budget) \\
Ada-KV \citep{adakv}         & KV & tok/hd    & pf    & \none & agn & free & evict (error bound) \\
Quest \citep{quest}          & KV & pg        & dec   & \full & qry & free & select (keep all) \\
KIVI \citep{kivi}            & KV & bit       & dec   & \half & agn & free & quantize \\
Palu \citep{palu}            & KV & rk        & pf    & \half & agn & adpt & low-rank \\
MLA \citep{mla}              & KV & rk        & pre   & \half & agn & scr  & low-rank (latent) \\
CaM \citep{cam}              & KV & tok       & dec   & \none & agn & free & merge \\
\midrule
LLMLingua \citep{llmlingua}      & Prompt & span & pf & \none & agn & free & prune (perplexity) \\
LongLLMLingua \citep{longllmlingua} & Prompt & span & pf & \none & qry & free & prune (query-aware) \\
Gisting \citep{gisting}      & Prompt & gist  & pf    & \none & agn & adpt & encode-to-latent \\
ICAE \citep{icae}            & Prompt & gist  & pf    & \none & agn & adpt & autoencode \\
xRAG \citep{xrag}            & Prompt & gist  & pf    & \none & agn & adpt & embed-as-token \\
Cartridges \citep{cartridges}& Prompt & param & x-task& \half & agn & adpt & offline distill \\
\midrule
RMT \citep{rmt}              & Arch & st       & pre   & \none & agn & scr & segment recurrence \\
Infini-attn \citep{infiniattention} & Arch & st & pre  & \none & agn & scr & compressive memory \\
Mamba \citep{mamba}          & Arch & st       & pre   & \none & agn & scr & SSM constant state \\
Titans \citep{titans}        & Arch & st       & pre   & \half & qry & scr & test-time memory \\
DMC \citep{dmc}              & Arch & tok      & pre   & \none & lrn & adpt & learned accumulate \\
\midrule
MemGPT \citep{memgpt}        & Agent & fact    & task  & \full & qry & ctrl & page / archive \\
ACON \citep{acon}            & Agent & span    & task  & \none & agn & ctrl & guideline summarize \\
MEM1 \citep{memone}          & Agent & span    & task  & \none & lrn & RL   & learned consolidate \\
Mem0 \citep{mem0}            & Agent & fact    & x-task& \half & qry & ctrl & add/update/delete \\
MemoryBank \citep{memorybank}& Agent & fact    & x-task& \half & agn & ctrl & decay curve \\
RAPTOR \citep{raptor}        & Agent & fact    & x-task& \full & qry & ctrl & tree summary + retrieve \\
\midrule
NSA \citep{nsa}              & Sparse & pg      & pre   & \half & qry & scr  & learned sparse attn \\
MoBA \citep{moba}            & Sparse & pg      & pre   & \half & qry & scr  & block routing \\
DuoAttention \citep{duoattention} & Sparse & hd & dec  & \half & agn & adpt & retrieval/stream heads \\
MInference \citep{minference}& Sparse & pg      & pf    & \half & qry & free & dynamic prefill sparsity \\
\midrule
ToMe \citep{tome}            & MM & vtok        & pf    & \none & agn & free & merge visual tokens \\
LOOK-M \citep{lookm}         & MM & tok/vtok    & pf    & \none & agn & free & evict + merge \\
PagedAttention \citep{pagedattention} & Sys & tok & srv & \full & agn & free & paged reuse \\
InfiniGen \citep{infinigen}  & Sys & tok        & srv   & \half & qry & free & offload + select \\
\bottomrule
\end{tabular}
\caption{Representative methods classified along the seven axes of
Section~\ref{sec:taxonomy}. The table makes the survey's organizing claim concrete:
methods that share no venue or vocabulary nonetheless occupy nearby points in this
space, and the columns that vary most, namely granularity, lifecycle,
reversibility, and query-adaptivity, are exactly those the rate--distortion lens
predicts are decisive.}
\label{tab:master}
\end{table*}

\end{document}